\documentclass[acmsmall,onecolumn]{acmart}

\usepackage{algorithm}
\usepackage[noend]{algpseudocode}
\usepackage{graphicx}
\usepackage{textcomp}
\usepackage{xcolor}
\usepackage{soul}
\usepackage{textgreek}
\usepackage{caption}

\usepackage{tikz}
\usepackage{multirow}
\usepackage{textgreek}
\usepackage{wrapfig,lipsum,booktabs}
\usepackage{amsmath}
\usepackage{subfig}

\algdef{SE}[DOWHILE]{Do}{doWhile}{\algorithmicdo}[1]{\algorithmicwhile\ #1}%

\usepackage{xspace}
\newcommand{\name}{{\sc{Stash}}\xspace}

\setcopyright{none}
\renewcommand\footnotetextcopyrightpermission[1]{}
\newcommand*\circled[1]{\tikz[baseline=(char.base)]{
          \node[shape=circle,draw,fill=black,text=white,font=\bf,inner sep=0.5pt] (char)
            {\scriptsize#1};}}
            
\newcommand*\tcircled[1]{\tikz[baseline=(char.base)]{
            \node[shape=circle,draw,thick,font=\bf,inner sep=1pt] (char) 
            {\footnotesize#1};}}

\iftrue

\newcommand{\XY}[1]{{\color{brown}{\small{\bf [XY: #1]}}}}

\newcommand{\sm}[1]{{\color{blue}{\small{\bf [Subrata: #1]}}}}

\else

\newcommand{\XY}[1]{}

\newcommand{\sm}[1]{}

\newcommand{\PQ}[1]{}

\fi

\begin{document}

\title{Analysis of Distributed Deep Learning in the Cloud} 

\author{Aakash Sharma}
\affiliation{%
 \institution{The Pennsylvania State University}}
  \email{abs5688@psu.edu}
 \author{Vivek M. Bhasi}
\affiliation{%
 \institution{The Pennsylvania State University}}
  \email{vmb5204@psu.edu}
 \author{Sonali Singh}
\affiliation{%
 \institution{The Pennsylvania State University}}
  \email{sms821@psu.edu}
 \author{Rishabh Jain}
\affiliation{%
 \institution{The Pennsylvania State University}}
  \email{rishabh@psu.edu}
 \author{Jashwant Raj Gunasekaran}
\affiliation{%
 \institution{Adobe Research}}
  \email{jgunasekaran@adobe.com}
 \author{Subrata Mitra}
\affiliation{%
 \institution{Adobe Research}}
  \email{subrata.mitra@adobe.com}
 \author{Mahmut Taylan Kandemir}
\affiliation{%
 \institution{The Pennsylvania State University}}
  \email{mtk2@psu.edu}
 \author{George Kesidis}
\affiliation{%
 \institution{The Pennsylvania State University}}
  \email{gik2@psu.edu}
 \author{Chita R. Das}
\affiliation{%
 \institution{The Pennsylvania State University}}
 \email{cxd12@psu.edu}

\begin{abstract}
Deep neural networks (DNNs) are increasingly popular owing to their ability to solve complex problems such as image recognition, autonomous driving, and natural language processing. Their growing complexity coupled with the use of larger volumes of training data (to achieve acceptable accuracy) has warranted the use of GPUs and other accelerators. Such accelerators are typically expensive, with users having to pay a high upfront cost to acquire them. For infrequent use, users can, instead, leverage the public cloud to mitigate the high acquisition cost.
However, with the wide diversity of hardware instances (particularly GPU instances) available in  public cloud, it becomes challenging for a user to make an appropriate choice from a cost/performance standpoint.
\par
In this work, we try to address this problem by (i) introducing a comprehensive
distributed deep learning (DDL) profiler \name, which determines the various execution stalls that DDL suffers from, and (ii) use \name to extensively characterize various public cloud GPU instances by running popular DNN models on them. Specifically, it estimates two types of communication stalls, namely, interconnect and network stalls, that play a dominant role in DDL execution time.
\name is implemented on top of prior work, DS-analyzer,  that computes only the CPU and disk stalls. Using our detailed stall characterization, we list the advantages and shortcomings of public cloud GPU instances for users to help them make an informed decision(s). Our characterization results indicate that the more expensive GPU instances may not be the most performant for all DNN models and that AWS can sometimes sub-optimally allocate hardware interconnect resources. Specifically, the intra-machine interconnect can introduce communication overheads of up to 90\%  of DNN training time and the  network-connected instances can suffer from up to 5$\times$ slowdown compared to training on a single instance. Furthermore, (iii) we also model the impact of DNN macroscopic features such as the number of layers and the number of gradients on communication stalls, and finally, (iv) we discuss network stall analysis at scale through an idealized scaling model and empirical observations.
\vspace{-2.75mm}
\end{abstract}

\maketitle
\pagestyle{plain}


\section{Introduction} \label{sections: introduction}
The continual growth of deep learning
has fuelled many facets of Artificial Intelligence such as machine vision \cite{image-transformer}, 
speech \cite{attention-layer}, autonomous driving \cite{DAD}, natural language processing \cite{googleBERT}, etc. The advancements in deep learning have mainly been driven by the availability of large amounts of training data as well as powerful compute platforms such as CPU or GPU clusters \cite{AWS-HPC, Azure-HPC}, TPUs \cite{tpu0}, NPUs \cite{npu0} and other accelerators that can handle increasingly complex/heavy Deep Neural Network (DNN) computations. However, the ever-growing DNN-model and training data sizes accompanied by the increasing ubiquity of DNNs place a higher demand on compute resources for faster processing speeds and shorter overall training time. Although current accelerators enable faster training, they are typically expensive to maintain, owing to their power-hungry nature. This potentially renders them cost-ineffective, especially in intermittent training scenarios. To avoid the prohibitively high upfront cost of purchasing a GPU machine/cluster, users employ public cloud GPU resources \cite{AWS-HPC,Azure-HPC} to run their workloads.\par

Public cloud providers such as AWS \cite{aws}, Azure \cite{azure}, and GCP \cite{gcloud}  provide a gamut of GPU instance offerings. These offerings vary in their hardware configurations and pricing. Cloud providers typically do not allow any flexibility in changing the CPU vCores, memory or GPUs of an instance, thereby limiting users to select from \emph{pre-configured} instances.
Note that the choice of instance type(s) drives the total cost of training a model \cite{dawnBench} and users may rely on benchmarks such as DawnBench \cite{dawnBench}, NVIDIA examples \cite{nvidia_pyTorch_egs}, etc. or on their intuition to choose the best type of instance(s) for their needs.
As we demonstrate in this paper, it is non-trivial to choose the most performant and/or cost-effective instance configuration for model training.

To address this problem, we introduce a Distributed Deep Learning (DDL) profiler, \name, which can measure the various execution stalls (on network, CPU and disk) that a typical DDL pipeline experiences.
Using our profiler, we characterize various public cloud GPU instances from both a cost and performance perspective with emphasis on  communication-related stalls.
This characterization provides novel insights into public cloud GPUs and its network, which can be used by tenants to make an informed decision vis-a-vis choosing the right DDL cluster configuration for their specific model. Our profiler \name, is built upon the DS-analyzer profiler \cite{ds-analyzer} released by Microsoft fiddle \cite{msr-fiddle}.
DS-analyzer, introduced by Mohan et al. \cite{mohan2021}, characterizes single-node DNN jobs in a private Microsoft cluster with emphasis on the bottlenecks (stalls) caused due to CPU pre-processing and storage I/O latency.\par

However, it has a key omission of not profiling communication-related stalls.
Compared to the single-node scenario, where the primary stalls were observed to be CPU and/or disk I/O stalls in the DS-Analyzer work, we observe communication stalls to be the primary bottleneck in both single and multi-node DDL (which is also corroborated by prior works \cite{pipedream, wang2019characterizing}).
As mentioned earlier, DS-Analyzer fails to account for this overhead, which can take up to 90\% of the training time (Section~\ref{sections: evaluation}, Fig~\ref{fig:p2_interconnect_stall}).
In fact, although storage-related stalls can (at least partially) be eliminated through DRAM caching in early epoch(s), communication stalls hamper every iteration of a typical DDL, thus proving to be a more pressing concern.
Apart from overlooking this, the study was conducted in a private data center with its own unique hardware and performance characteristics.
Low-level details of various devices such as hardware vendor/model, interconnect used etc. were not released, thereby severely diminishing its utility to users who do not have access to the specific cluster. 
Motivated by this, we build upon this work by (i) \emph{focusing on communication overheads incurred during DDL}, and (ii) \emph{characterizing publicly-available instances so that the work is beneficial to a larger audience.}\par

We propose novel techniques to profile the communication-related stalls of DDL and implement it as part of our profiler, \name. 
The profiler is fully automated and users can use it to find stalls in any DDL model (using data parallelism) by adding "profiling hooks" to their training loop.
Next, using this profiler, we extensively characterize public cloud GPU instances for the various stalls they suffer from while executing a typical DDL pipeline. A stall analysis on public cloud (AWS in our experiments) is particularly useful, since various instance types differ not just in the hardware they offer, but also in the QoS they provide, as discussed in later sections. Moreover, a systematic study of the communication overhead of public cloud instances for DDL is lacking, partly due to the lack of publicly-available tools or profiling methodologies to measure such an overhead.
We attempt to bridge this gap by carefully measuring the communication stalls (along with CPU and I/O stalls) on various GPU-accelerated instances of a public cloud during DDL. Furthermore, using this profiler, we analyze a number of DNN architectures to understand which architectural properties (such as the number and sizes of  layers) drive communication stall behavior.\\
\vspace{-0.5mm}
Our {\bf main contributions} in this paper can be summarized as follows:
\begin{itemize}
\item{We introduce \name, a profiling tool which can measure communication stalls (in addition to CPU and disk stalls) of DDL workloads running on both single and multiple nodes.}
\item{We perform stall analysis on various AWS GPU instances, using a number of popular DNN models. The estimated communication overheads from intra-machine interconnect are found to be up to 90\% of the training time and network-connected instances are found to be slowed down by up to 5$\times$ compared to a single node instance. Our profiling has led us to some surprising discoveries vis-a-vis the communication overhead experienced by AWS GPU instances. \emph{We list out the summary of our key findings in Table~\ref{tab: key_insights} at the end.}}
\item{We identify the limitations of each instance type. 
Specifically, our results indicate that higher 
capacity GPU instances do not always lead to better performance and that AWS hardware interconnects have various shortcomings. }
\item{We identify architectural features in DNN models that influence communication stall behavior, namely, the number of layers as well as the total number of parameters (size of the DNN model).}
\item{Finally, we discuss how network stalls vary when scaling instances and propose an idealized model for the empirical observation.}
\end{itemize}
The rest of this paper is organized as follows. 
In Section~\ref{sections: background}, we discuss the background pertinent to DDL, AWS GPU instances, and DS-Analyzer. In Section~\ref{sections: motivation}, we motivate our problem. Our characterization scheme is described in Section~\ref{sections: methodology}.
The results from our characterizations are presented in Sections~\ref{sections: evaluation} and~\ref{sections: evaluation_synthetic} along with the network stall scaling analysis.
In Section~\ref{sections: relatedwork}, we provide an overview of the related work in this area, and finally,  Section~\ref{sections: summary} summarizes our major observations and findings.
\vspace{-2mm}

\section{Background} \label{sections: background}


Several distributed training methods have been proposed to parallelize the computations on multiple machines. 
These distributed deep learning (DDL) techniques broadly fall under two categories: data parallel \cite{pytorch_dist} and model parallel \cite{model_pll} training. 
In the former, the model is replicated and data are  partitioned across multiple devices for parallel computation, whereas in the latter, the model is split into multiple partitions, which are processed on different GPU devices. 
We limit the scope of this work to studying data parallel DDL, since extending \name to model parallelism would be a work in its own right; we postpone it to a future study.
In this section, we summarize the main steps involved in data parallel training and then provide an overview of the different public cloud offerings of AWS for deep learning.
\vspace{-1.5mm}
\subsection{Data-Parallel Distributed Deep Learning} \label{subsection: ddl_background}
\begin{figure}[!tbp]
  \centering
  \begin{minipage}[b]{0.49\textwidth}
    \includegraphics[width=\textwidth]{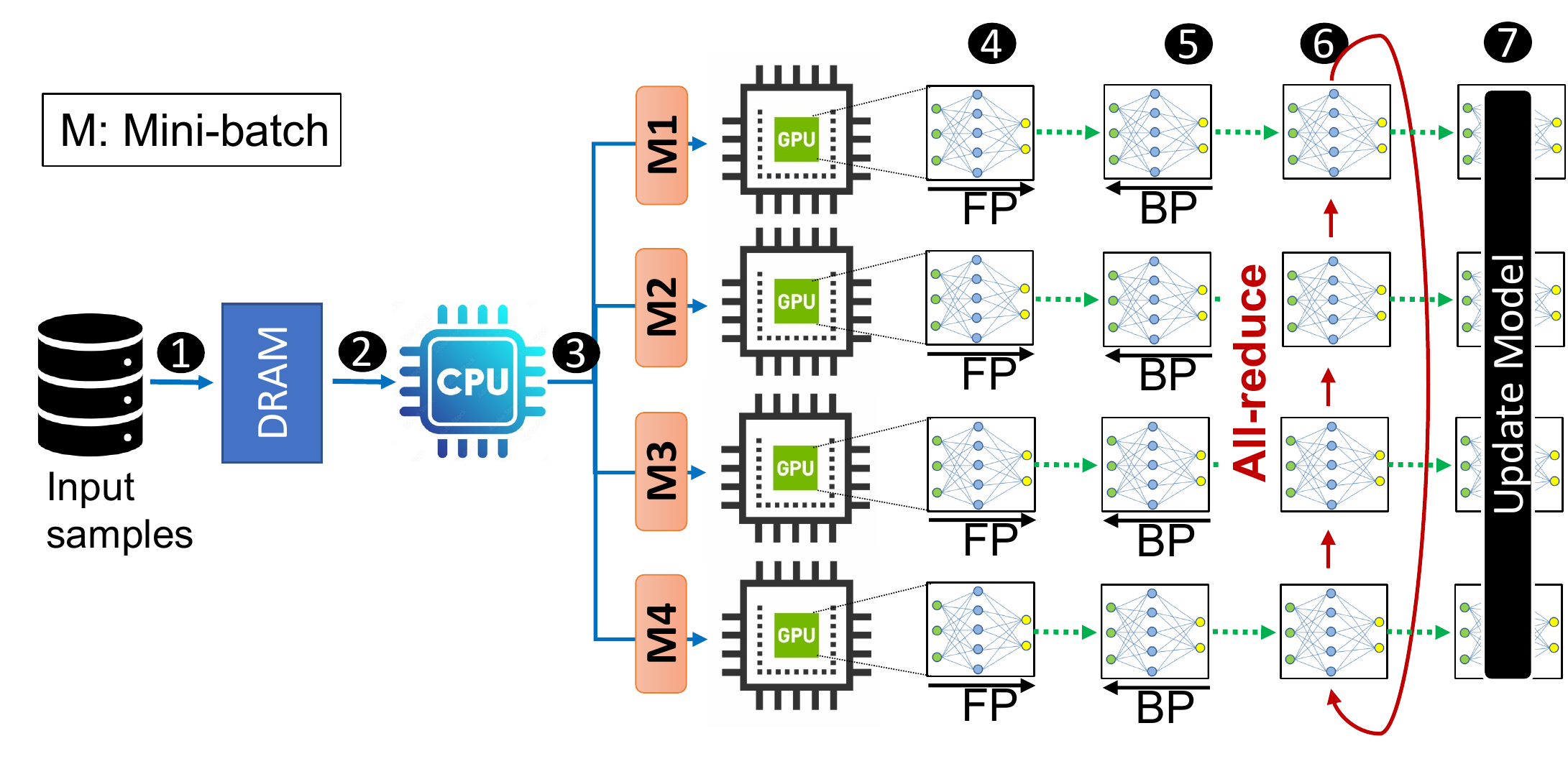}
    \caption{A typical Distributed DNN training pipeline.}
    \vspace{-4mm}
    \label{fig: dnn_train}
  \end{minipage}
  \hfill
  \begin{minipage}[b]{0.49\textwidth}
    \includegraphics[width=\textwidth]{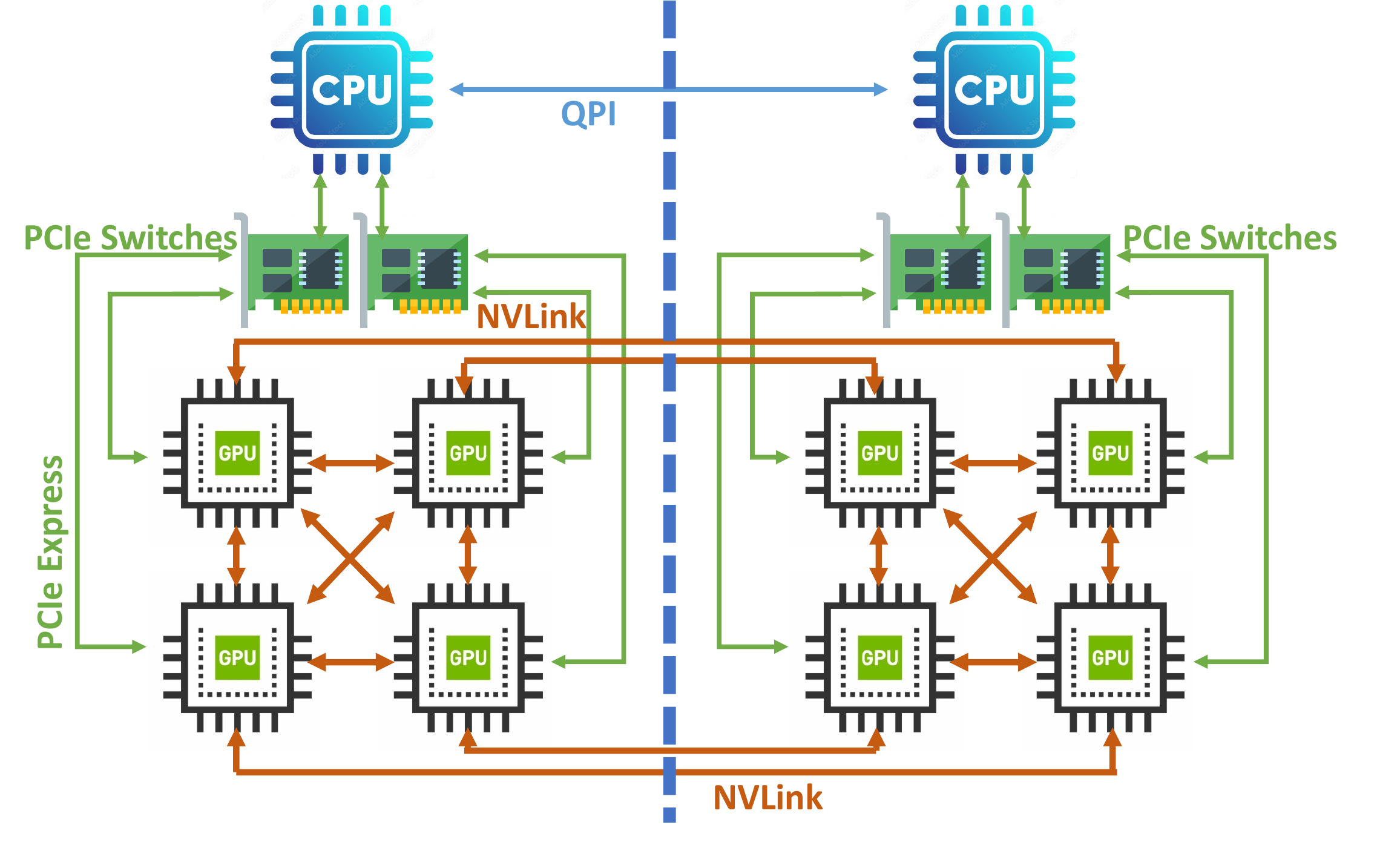}
    \caption{P3.16xlarge interconnect architecture.}
    \vspace{-4mm}
    \label{fig:p3_interconnect}
  \end{minipage}
  \vspace{-1.5mm}
\end{figure}

Figure \ref{fig: dnn_train} outlines the typical process of distributed DNN training
(deep learning) where several epochs comprising multiple iterations are performed. For each of these iterations, a mini-batch of data items (training samples) is first fetched from storage \circled{1} (and cached in DRAM). These data items are then pre-processed to make them amenable to training \circled{2}. To perform training, the data items are first partitioned among all workers (GPUs) where each holds a copy of the DNN model \circled{3}. Each worker then performs a forward pass (FP) to obtain the model's current prediction on its partition and a loss value is calculated depending on its deviation from the ground truth \circled{4}. 
Next, each worker performs a backward pass (BP) to compute the gradients of the loss function with respect to the model parameters (weights) \circled{5}. This is followed by a synchronization step  wherein the weight gradients are exchanged (communicated) and aggregated across all workers via a collective all-reduce operation \circled{6}.  Here, given that there are $n$ workers and gradients totalling $G$ bytes, each worker sends its fraction of gradients, $G/n$, over the network $n-1$ times for each of the two phases of an idealized all-reduce, namely, share-reduce and share-only. 
Therefore, the total data sent by a worker across the network is $2(G/n)(n-1)$ \cite{all-reduce-towards-data-science}. 
Finally, the aggregated weight gradients from all-reduce are used to calculate the optimized weight updates which are then applied to all models in parallel \circled{7}. After this, the same steps are repeated for the next iteration until the completion of the epoch. Note that, for the next epoch, training samples may be fetched from DRAM due to caching in \circled{1}. 

In the above procedure, the different steps are primarily associated with specific hardware components. For instance, step \circled{1} involves storage I/O, step \circled{2} is typically CPU-bound, while steps \circled{3}, \circled{4}, \circled{5}, \circled{6} and \circled{7} are GPU-bound. 
Additionally, while performing the weight gradient aggregation \circled{6} and sometimes even during BP \circled{5}, the gradients are communicated amongst the workers either using interconnects (such as PCIe and NVLink) and/or network links (such as Ethernet and Infiniband).


\vspace{-1.5mm}
\subsection{Public Cloud Offerings} \label{subsection: public_cloud_offerings}
Hardware capabilities, both in terms of compute and interconnects, are particularly significant in the context of DDL on the public cloud, as the GPU instances offered by providers (such as AWS) have fixed configurations \cite{aws_nvidia}, thereby limiting user choice of a custom single-node training solution. 
Table~\ref{tab: public cloud gpu offerings} lists the P family GPU instance types offered by AWS along with their hardware specifications and pricing.
The P4 instances have the most powerful GPUs (NVIDIA A100 Tensor core GPUs), while the P3 and P2 instances respectively have the less powerful, yet quite capable, NVIDIA V100 and K80 GPUs. The P3 and P2 instances are of particular interest to us, as they offer the most variety in terms of the number of GPUs available per node amongst all GPU instances viable for DNN training.\footnote{P4 is a dedicated offering not considered herein.}

Apart from the GPUs used, the interconnects and network links available to these instance types also have a significant impact on the end-to-end training time as they dictate data transfer speeds during various training steps. Specifically, interconnect links are utilized during gradient communication among workers (GPUs) present on the same physical node,  whereas network links are used when the communication is between workers on different nodes. The interconnect architecture for the AWS p3.16xlarge and p3.24xlarge instances \cite{p3_interconnect} is depicted in Figure \ref{fig:p3_interconnect}. 

We aim to demonstrate in this work that choosing the optimal instance configuration from these offerings to maximize performance, while minimizing cost, is a non-trivial problem as various factors, including GPU type, core count, number of instances, interconnects/network links used, and DNN model type can all play crucial and complex roles in determining this.  Furthermore, we perform characterization on possible instance configurations (distributed and single-node) to identify training bottlenecks. 

\begin{table}[!ht]
    \centering
    \scriptsize
    \begin{tabular}{|c|c|c|c|c|c|c|c|c|}
    \hline
        \multicolumn{2}{|c|}{\textbf{Instance type(s)}} &  \textbf{GPU(s)} & \textbf{VCPUs} & \textbf{Interconnect} & \textbf{GPU Memory (GB)} & \textbf{Main Memory (GB)} & \begin{tabular}[c]{@{}l@{}}\textbf{Network} \\ \textbf{Bandwidth}\\ \textbf{(Gbps)}\end{tabular} & \textbf{Price/hr}\\ \hline
        \multicolumn{2}{|c|}{P4} & 8$\times$A100 & 96 & NVSwitch & 320 & 1152 & 400 & \$32.7726\\ \hline
        \multirow{4}{*}{P3} & p3.2xlarge & 1$\times$V100 & 8 & PCIe & 16 & 61 & up to 10  & \$3.06 \\ \cline{2-9}
        ~ & p3.8xlarge & 4$\times$V100 &  32 & PCIe + NVLink & 64 & 244 & 10 & \$12.24\\ \cline{2-9}
        ~ & p3.16xlarge & 8$\times$V100 & 64 & PCIe + NVLink & 128 & 488 & 25 & \$24.48\\ \cline{2-9}
        ~ & p3.24xlarge & 8$\times$V100 & 96 & PCIe + NVLink & 256 & 768 & 100 &  \$31.218\\ \hline
        \multirow{3}{*}{P2} & p2.xlarge & 1$\times$K80 & 4 & PCIe & 12 & 61 & < 10 & \$0.90\\ \cline{2-9}
        ~ & p2.8xlarge & 8$\times$K80 & 32 & PCIe & 96 & 488 & 10 & \$7.20\\ \cline{2-9}
        ~ & p2.16xlarge & 16$\times$K80 & 64 & PCIe & 192 & 732 & 25 & \$14.40\\ \hline
    \end{tabular}
\caption{AWS GPU instance types with prices (N. Virginia).}
\vspace{-10mm}
\label{tab: public cloud gpu offerings}
\end{table}

\subsection{DS-Analyzer: Stall Characterization} 
Among the works that characterize the private cloud \cite{hu2021characterization, mojumder2018profiling, jain2019performance, awan2017depth,awan2019scalable,ko2021depth}, DS-Analyzer\cite{mohan2021} is of particular significance to us as it also aims to identify DNN training bottlenecks, specifically with regards to `fetch' and `prep' stalls.
These stalls refer to the time spent fetching mini-batches of data from the disk (fetch stall) and pre-processing it prior to training with it (prep stall) in a deep learning iteration. 

DS-Analyzer uses three steps to calculate prep and fetch stalls (refer to Figure \ref{fig: stash_scheme}). Step~\tcircled{2} pre-populates synthetic data in the GPUs and runs training to measure the maximum ingestion rate of the system. This is followed by step~\tcircled{3} which runs training on actual data but with all OS caches cleared\footnote{In the original paper, this step is described to be after the next step but we observed from the DS-analyzer open source code that that is not the case.}. Finally, in step~\tcircled{4}, training is run over actual data such that the entire data is cached in main memory (from the previous step). The prep stall is calculated by finding the difference between \tcircled{4} and \tcircled{2}. This is because there is no disk I/O involved in step \tcircled{4} and any difference in training time after deducting the time spent in GPU processing of \tcircled{2} will yield the time spent in pre-processing at CPU. After this, the fetch stall is calculated by finding the difference of \tcircled{3} and \tcircled{4}, since any increase in time over \tcircled{4} would be due to disk I/O.\par  

\section{DDL Stall Analysis in Public Cloud} \label{sections: motivation} 
In this section, we highlight the novelty of this work over prior work. We also motivate the importance of "profiling communication stalls" in DDL, especially on the public cloud.

\noindent\textbf{What are the limitations of prior work?} \\
As stated earlier, \cite{ds-analyzer} profiles DDL jobs in a private Microsoft cluster for CPU and disk stalls.
However, it fails to account for the most dominant and consequential of all stalls in distributed training, namely, communication stalls.
While disk stalls can (at least partially) be eliminated through caching in DRAM (as part of data fetch in the first epoch), communication stalls will remain a part of every iteration of the DDL.
And, although frameworks such as PyTorch distributed \cite{pytorch_dist} overlap communication with computation, stalls due to communication remain dominant as shown in our evaluations (Section~\ref{sections: evaluation}).
Hence, the lack of communication stalls in the prior work severely limits its usability.\par
Moreover, in this case, the characterization is conducted over a private cluster with little hardware architecture details available. While readers can learn about data stalls in DDL from this work, it is of little use to general users who run their workloads on hardware different from that used in the study.
On the other hand, characterizing public cloud instances has higher utility to a broader range of users as the hardware is publicly available. Additionally, 
 we identify specific macroscopic DNN architecture --specific features that contribute to communication stalls-- as opposed to prior work that fails to do so.

\noindent\textbf{Are there prior works which analyze communication overhead in DDL?} \\
In \cite{pipedream}, the authors measure the communication overhead of training to be 80\% of the entire training time.
However, they do not specify any general methodology to measure the actual overhead. In comparison, 
\cite{wang2019characterizing} characterizes DDL workloads on Alibaba PAI \cite{alibaba_pai} and observes the communication overhead to be 62\% with the use of parameter server (PS) \cite{mosharraf_parm_server} (whose communication performance is strictly less than all-reduce \cite{wang2019characterizing,li2019evaluating}).
Moreover, the said work is specific to a private cluster and specific hardware details of the machines running the workload are not available. Although they have released a general profiling methodology using TensorFlow~\cite{tensorflow} internal tooling and manual feature extraction, it is limited to TensorFlow and the PS communication architecture.
Finally, \cite{li2020characterizing} characterizes DDL training on Google Cloud. They build a general framework for measuring the DDL performance, specifically for transient cloud instances which are frequently revoked. This work, too, does not account for communication stalls.  The remaining related works are discussed later in Section~\ref{sections: relatedwork}.

\noindent\textbf{Why should a communication profiler for DDL be introduced?}\\
A communication stall profiler can potentially help in the end goal of reducing it.
In the past, several distributed DNN algorithms have been proposed  \cite{zhang2017poseidon,zhang2015deep,lian2018asynchronous,mosharraf_parm_server,ho_more_effective_distributed_ml,goyal2017accurate,jeff_dean_Large_Scale_Distributed_Deep_Networks} to reduce communication overhead of DDL. However, there is a lack of a {\em profiling tool} to measure the specific efficacy with regards to the communication overhead for various algorithms. 
Even when the efficacy study exists for specific algorithms, ML scientists lack the tools to measure the various communication overheads that their optimizations may suffer from. An ML scientist may remain oblivious to the slowdown caused by communication links on their otherwise high-performing algorithm with a single GPU.

\noindent\textbf{Why characterize GPU instances of public clouds?}\\
The gamut of public cloud GPU-based instances
available (see Section~\ref{subsection: public_cloud_offerings}) makes the task of choosing the most performant configuration a non-trivial one for end-users. 
This is due to the presence of various stalls in the DDL pipeline as well as the lack of a good scientific study that characterizes these stalls in the various public cloud GPU instances. Prior works on characterizing deep learning such as \cite{hu2021characterization, mojumder2018profiling, jain2019performance, awan2017depth,awan2019scalable,ko2021depth,mohan2021,li2020characterizing,wang2019characterizing} do not characterize the  instances of the public cloud for their QoS; consequently,  users cannot use these works to choose the appropriate instance type(s). To further complicate matters, cloud providers offer different types of interconnects for their GPU-accelerated instance type(s). Apart from the interconnect type, the user can also "tie" various instances through a computer network. These communication options introduce further complexity to the choice of instance(s) for DDL. In the absence of a characterization and profiling technique for choosing instance type(s), users can only guess the best instance to run their ML workloads based on their intuition and/or marketing from the cloud provider.

\noindent\textbf{How to lower DDL cost on public cloud?}\\
For a typical user, the cost of running training is an important metric to consider when using the public cloud, i.e., "cloud-spend" \cite{cloudspend}. Note that this is, in turn, primarily determined by the choice of public cloud instance(s). In this work, we try to answer this question by characterizing various AWS P instance types with regards to their "stall behavior" during DDL. We also identify the features in the DNN architecture which impact the communication stalls observed in training.

\section{Methodology} \label{sections: methodology}




    

\begin{figure}
  \centering
  \includegraphics[width=0.3\textwidth]{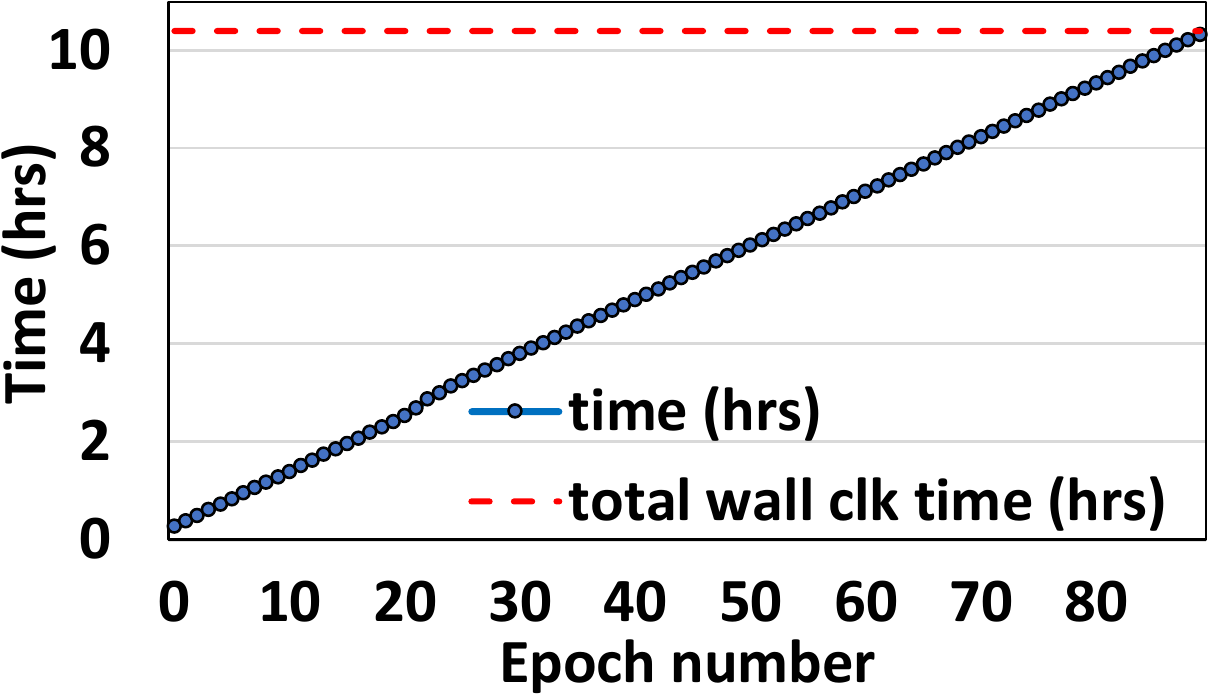}
  \caption{End-to-end ResNet18 training time on ImageNet vs. number of epochs. Training time scales linearly with \#epochs.}
  \label{fig: train_resnet}
\end{figure}

In this work, we use the AWS public cloud to run all our experiments. Specifically, we run DDL experiments  in the AWS N. Virginia region using P type instances, which are AWS's recommended instance type for performing deep learning. All DDL models are run using PyTorch distributed \cite{pytorch_dist} with data parallelism. Using our profiler \name, we characterized various AWS P type instances with reference to four stall parameters, namely, (i) interconnect stall,  (ii) network stall, (iii) CPU stall (prep stall), and (iv) disk stall (fetch stall). While these stalls provide important insights into the hardware characteristics of AWS instances, we also provide a training time and monetary cost comparison of running DDL on various AWS instance types. Our characterization exploits the repetitive nature of deep learning,  and is able to calculate the various stalls from a single epoch. This is possible since the stall characteristics of a single epoch are representative of that of the entire training time (which scales linearly with the number of epochs). To verify this, consider the plot of training time versus epoch number during end-to-end training of a ResNet18 DNN \cite{resnet} on the ImageNet dataset \cite{imagenet_ILSVRC2012} in  Figure \ref{fig: train_resnet}. The model was trained for 90 epochs with a batch size of 1024 in a distributed data parallel regime on a cluster of 4 NVIDIA A100 GPUs. Upon the conclusion of training, the top-1 and top-5 validation accuracies were, respectively, 67.5\% and 87.8\%. As can be seen from the figure, the training time increases linearly with the number of epochs and the total training time (obtained as a sum of per-epoch training time over 90 epochs) matches the wall-clock time of training. 

\subsection{Characterization}
We conduct two types of characterizations, macro and micro, on AWS instances as explained below.\\

\noindent\textbf{\textit{Macro Characterization: }} We run DDL using vision models and an NLP model (transformer-based) to characterize the relevant GPU instances.
The models used in our experiments are listed in Table~\ref{tab: models_desc}.
We bucket the models by their gradient size into two types -- small and large.
\begin{table*}[ht]
\vspace{-2mm}
\scriptsize
\begin{tabular}{|l|l|l|l|l|}
\hline
\textbf{Domain} &
\textbf{Type}                                                           & \textbf{Name}                                     & \textbf{\begin{tabular}[c]{@{}l@{}}Model/Gradient size\\ (\#Parameters)\end{tabular}} & \textbf{\begin{tabular}[c]{@{}l@{}}Input Dataset\end{tabular}} \\ \hline
\multirow{7}{*}{\begin{tabular}[c]{@{}l@{}}Vision\end{tabular}} &
\multirow{5}{*}{\begin{tabular}[c]{@{}l@{}}Small\end{tabular}}  & AlexNet  \cite{alexnet}           & 9.63M            & \multirow{7}{*}{Imagenet 1k \cite{imagenet1k} (133 GB)}                                     \\ \cline{3-4}
&                                                                        & MobileNet-v2  \cite{mobilenet}    & 3.4M            &                                                                  \\ \cline{3-4}
                         &            & SqueezeNet  \cite{squeezenet}     & 0.73M    &                                                                  \\ \cline{3-4}
                                                                      &   & ShuffleNet  \cite{shufflenet_v2} & 1.8M    &                                                                  \\ \cline{3-4}
                                                                     &    & ResNet18  \cite{resnet}           & 11.18M   &                                                                  \\ \cline{2-4}
  & \multirow{3}{*}{\begin{tabular}[c]{@{}l@{}}Large\end{tabular}}  & ResNet50 \cite{resnet}           & 23.59M  &                                                                  \\ \cline{3-4}
                                                                         &   & VGG11 \cite{vgg}                 & 132.8M    &                                                                  \\ \cline{1-1}\cline{3-5}
 NLP &                                                                      & BERT-large \cite{bert}           & 345M     & SQuAD2.0 \cite{squad} (45 MB)                           \\ \hline
\end{tabular}
\caption{DDL models used.}
\label{tab: models_desc}
\vspace{-8mm}
\end{table*}

\noindent\textbf{\textit{Micro Characterization: }} We conduct micro characterization by running synthetic training experiments using two models -- ResNet and VGG.  As part of this, we study various aspects of the model architecture that influence the communication stalls including the number of layers. We also modulate certain model architecture features (such as "residual" network branches and batch normalization layers) to study their impact on communication stalls. 



\begin{figure*}[h]
    \centering
    \subfloat[High level view of the \name profiler]{{\includegraphics[width=.5\linewidth]{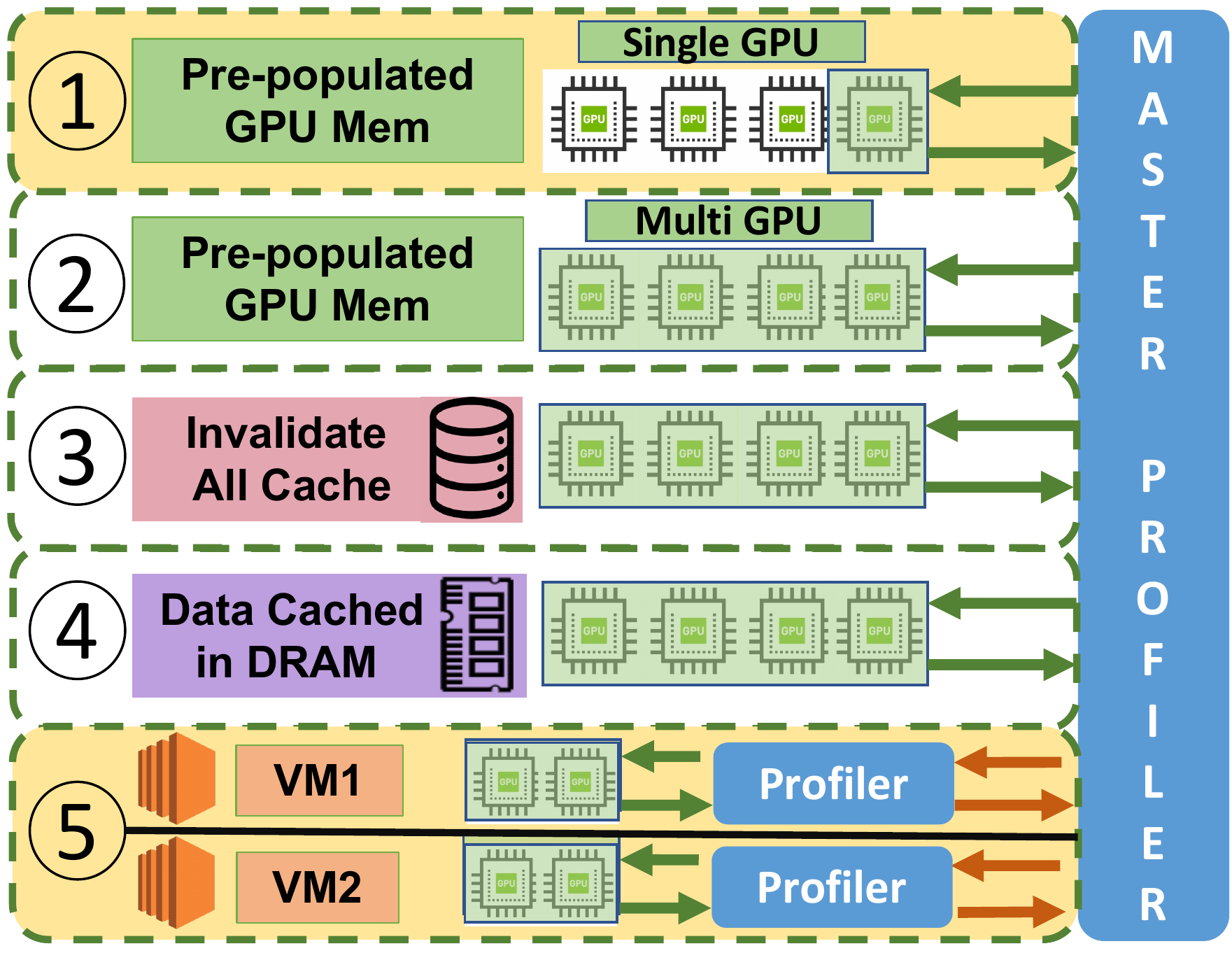} }}
    \subfloat[Interconnect stall calculation (M: Mini-batches, batch size = $\frac{n/4}{4}$]{{\includegraphics[width=.5\linewidth]{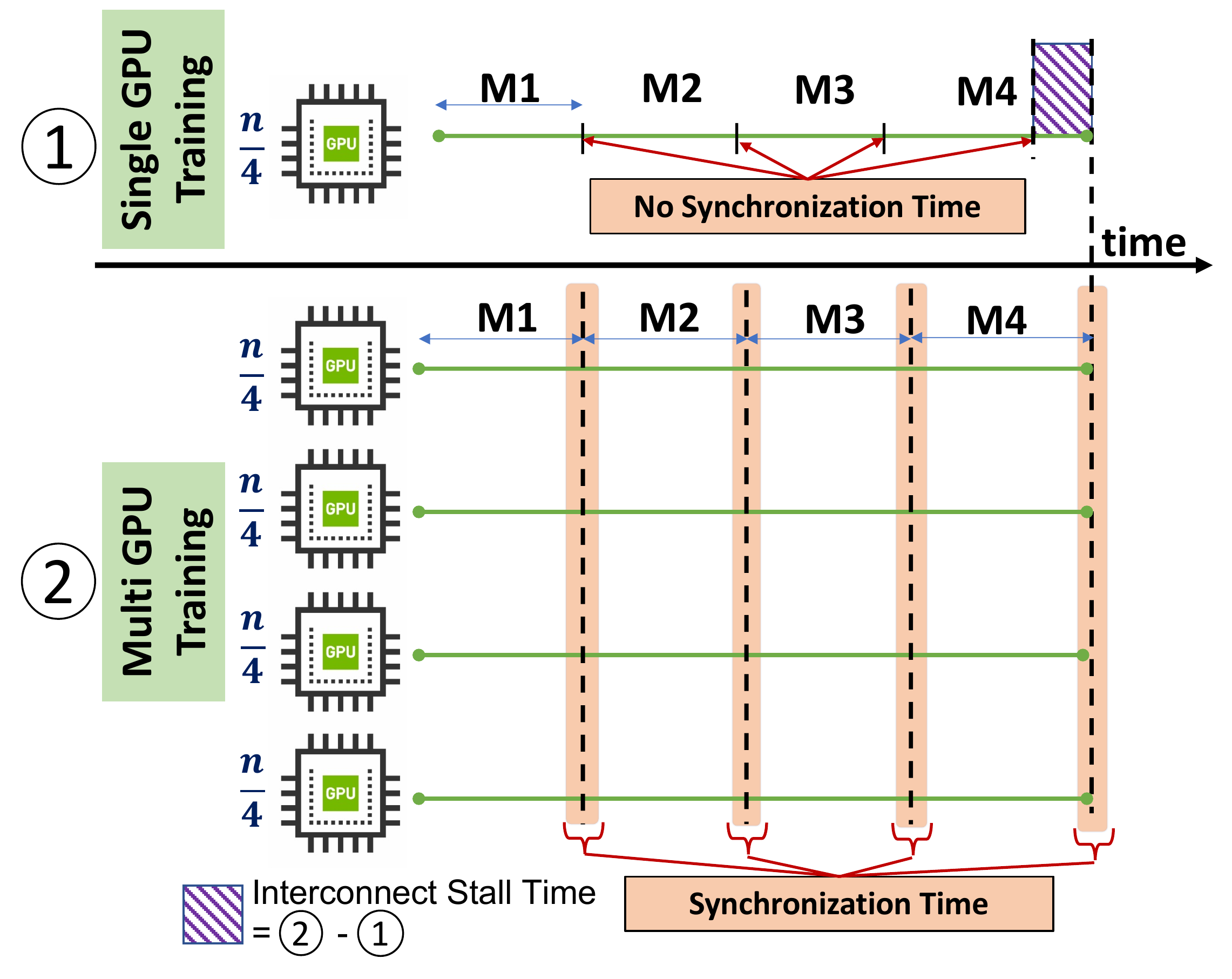} }}
    \vspace{-2mm}
    \caption{Our Stash scheme.}
    \label{fig: stash_scheme}
    \vspace{-5mm}
\end{figure*}


\vspace{-3mm}
\subsection{Profiler Design}
A schematic view of the \name profiler is depicted in Figure \ref{fig: stash_scheme}(a). As mentioned earlier, \name is an extension of the DS-Analyzer profiler. In the figure, step~\tcircled{1} and step~\tcircled{5} have been added to calculate the communication stalls and steps~\tcircled{2}, \tcircled{3}, and \tcircled{4} are from the prior work, DS-Analyzer. Note that \name pre-populates the GPU memory with synthetic training data as part of step~\tcircled{1}, \tcircled{2} and \tcircled{5}, and runs training on it. 
Training over the synthetically pre-populated data has the unique advantage of eliminating all CPU and disk stalls as neither disk I/O nor CPU pre-processing are happening while training. However, since training is run using multiple GPU workers, it is still bound to suffer from the communication stalls. Below, we describe the methodology of determining the communication stalls using step~\tcircled{1},  step~\tcircled{2}, and step~\tcircled{5}.

\subsubsection{\textbf{Interconnect Stall}}
We define an interconnect stall as the \textit{inter-GPU communication overhead of DDL in a single machine that arises due to the communication among the GPUs}. This is a key indicator of the performance of the underlying interconnect and is also indicative of the end-to-end training time. We determine the interconnect stall in two steps:  
\begin{enumerate}
\item \name pre-populates synthetic data in the memory of a single GPU only such that the number of samples the GPU processes is the same as that in a single GPU in a distributed training setup with multiple GPUs. Here, the batch size for multi-GPU training is kept the same as that of a single GPU. \name then performs synthetic training on just a single GPU (in a multi-GPU machine) while keeping all other GPUs idle (see step~\tcircled{1} in Figure \ref{fig: stash_scheme}(a)).
Since this is a single-GPU training, no inter-GPU communication overhead is incurred. 
\item \name then runs distributed training, over all GPUs in the machine, on synthetic data. The number of samples each GPU processes and the per-GPU batch size is kept the same as in step~\tcircled{1}. 
\end{enumerate}
Note that distributed training adds communication overhead to the end-to-end training time as a consequence of gradient synchronization. As a result, the difference in training time between \tcircled{2} and \tcircled{1} essentially yields the interconnect stall of the model for a particular machine. 


We now describe an example of determining interconnect stalls using Figure \ref{fig: stash_scheme}(b). Suppose that, in a four GPU machine, the total DNN training dataset consists of $n$ samples and the training must run over four mini-batches per epoch such that the batch size per GPU is set to be $\frac{n/4}{4}$. Therefore, as part of \tcircled{1}, \name will pre-populate only one GPU with $n/4$ samples and a training process will be launched for one epoch using that particular GPU, keeping the other GPUs idle. This single-GPU training epoch has no need for gradient synchronization and hence, it does not suffer from any communication overhead as shown in the figure. After step~\tcircled{1}, \name will pre-populate all other GPUs with $n/4$ samples each as part of step~\tcircled{2}, and launch distributed training 
over $n$ samples (i.e., a DDL epoch). These four training processes will suffer from communication overheads due to the all-reduce (gradient synchronization) step, as depicted in the figure.  The difference between the elapsed time of training over $n$ samples with 4 GPUs and training over $n/4$ samples with a single GPU is the "communication overhead" (indicated in figure), which is essentially the interconnect stall.


\subsubsection{\textbf{Network Stall}} \label{subsection: network_stall}
We define a network stall as the \textit{inter-GPU communication overhead of DNN training over multiple machines that arises due to the network link(s) between them}. This type of stall occurs when DDL is performed across multiple machines linked through a network. 
Since the all-reduce step requires gradients to be sent via both the intra-machine interconnect network as well as the inter-machine computer network, the slowest link becomes a  communication bottleneck. Whenever the network link is the slowest link (compared to intra-node interconnect), network stalls occur. 
\name determines network stalls as follows.
Synthetic distributed training is run over multiple machines connected via network such that the total number of GPUs is the same as in the single machine training of \tcircled{2}. This is step~\tcircled{5} in Figure \ref{fig: stash_scheme}. The difference between the training times of \tcircled{2} and \tcircled{5} yields the \textit{network stall} of the model.

\begin{figure}
  \centering
  \includegraphics[width=0.5\textwidth]{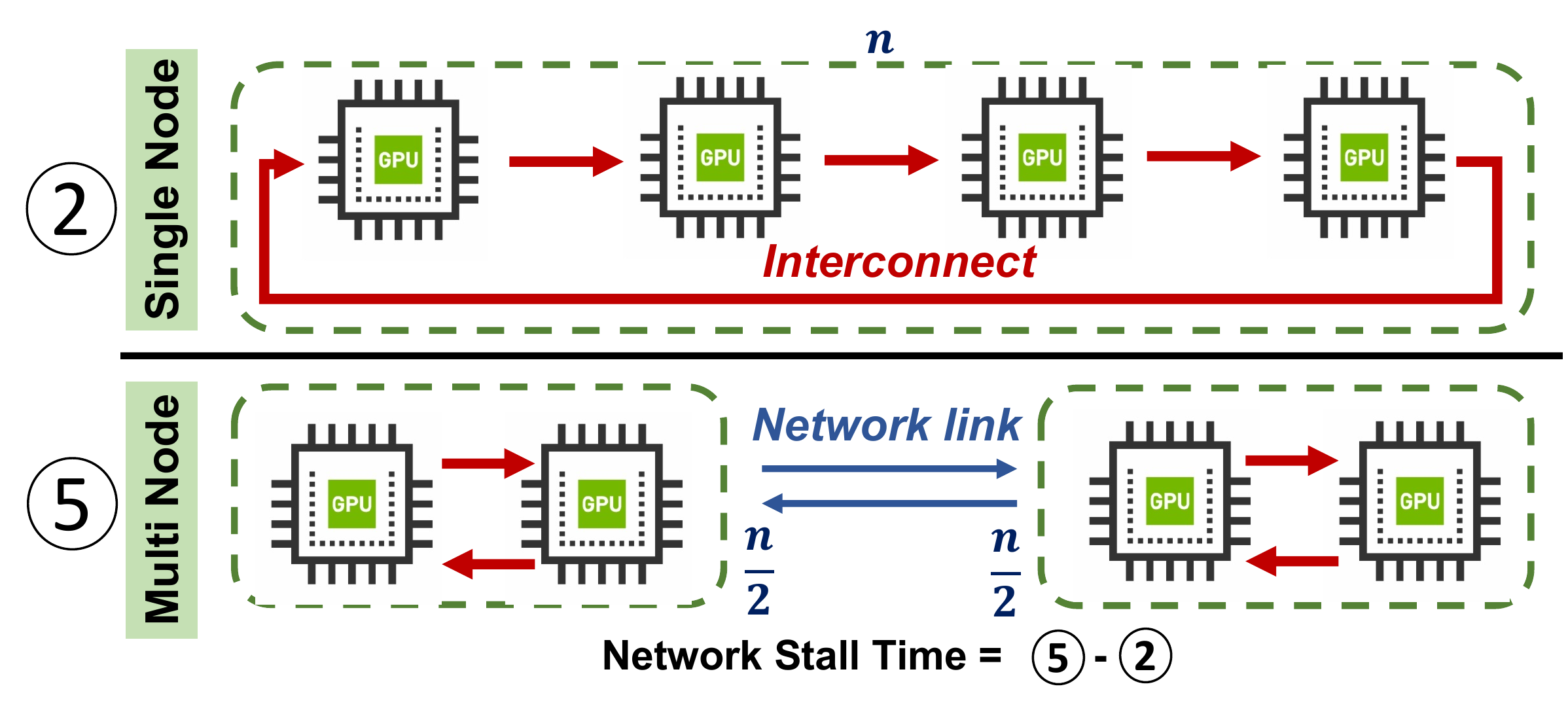}
  \caption{Network stall calculation.}
  \label{fig: nw_method}
  \vspace{-4mm}
\end{figure}

Again, we describe an example of determining a network stall using Figure \ref{fig: nw_method}. Suppose we run step~\tcircled{2} in an instance with 4 GPUs over $n$ data samples, as shown in the figure.  As part of \tcircled{5}, \name now runs training over 2 instances with 2 GPUs each but with $n/2$ samples per machine keeping the per GPU batch size constant.
When we train on 2 instances with a network link between them, the communication is bottlenecked by the network link if the link is slower than the hardware interconnect (most cases). 
For most cases where the network link is slower than the hardware interconnect, the network stall of the system is calculated as the time difference between \tcircled{2} and \tcircled{5}.\par

\section{Macro Characterization} \label{sections: evaluation}


Our characterization aims to answer a fundamental question, i.e.,  \textbf{which instance type is the most cost-effective and/or performant?} To answer this, we realize that further questions need to be asked and hence, we begin our discussion by asking a simple but specific question: \textit{How much stall does a DDL job experience from spending time on CPU, disk, interconnect, and network?}
We investigate this problem by conducting a stall analysis on AWS P type instances with representative DDL workloads, while keeping the GPU as the first class resource. 
Although we use specific DNN models as example workloads, the techniques used herein can be generalized to all DDL workloads.
We run our DDL workloads across four different mini-batch sizes (except BERT-large), with the largest batch size being (approximately) the maximum size that could fit in the GPU memory. For BERT, we only run on batch size 4, as that is the maximum size that allows the resultant data to fit in GPU memory (16 GB).
Note that the batch sizes stated in the figures are per GPU and the effective batch size can be obtained as the per-GPU batch size times the number of GPUs. For brevity, we only show the plots of the profiling with the smallest and largest batch sizes used. The stall percentage is calculated as: $I/C \enspace stall \% = (\frac{I/C \enspace stall \enspace  time}{single \enspace   GPU \enspace time}) \times 100$,$\quad$$N/W \enspace stall \% = (\frac{N/W \enspace stall \enspace  time}{single \enspace   instance \enspace time}) \times 100$, where the I/C and N/W stalls are calculated as described in the previous section. Our discussion starts in the sequel.
\begin{figure*}[h]
    \vspace{-4mm}
    \centering
    \subfloat[CPU stall \% (CPUs are sufficient for pre-processing)]{{\includegraphics[width=0.5\linewidth]{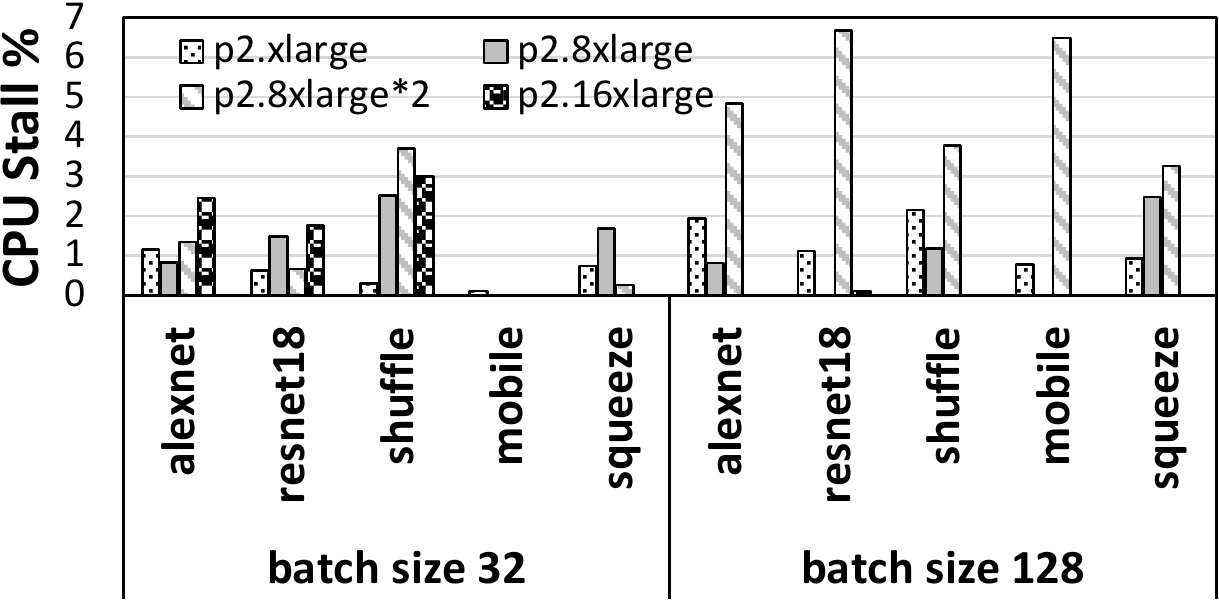} }}
    \subfloat[Disk stall \% (Scales with \#GPUs per instance.)]{{\includegraphics[width=0.5\linewidth]{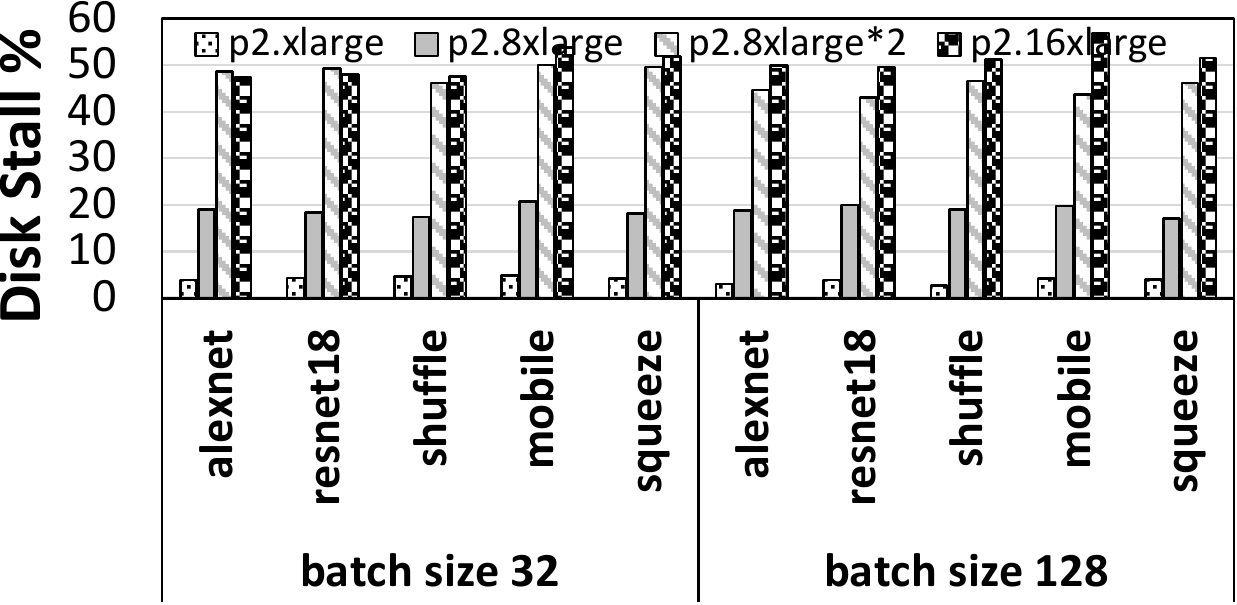} }}
    \vspace{-3mm}
    \caption{CPU and disk stall \% of total training time in P2.}
    \label{fig:p2_cpu_disk_stall}
\end{figure*}

\begin{figure*}
\vspace{-5mm}
\centering
    \subfloat[Interconnect stall time]{{\includegraphics[width=0.5\linewidth]{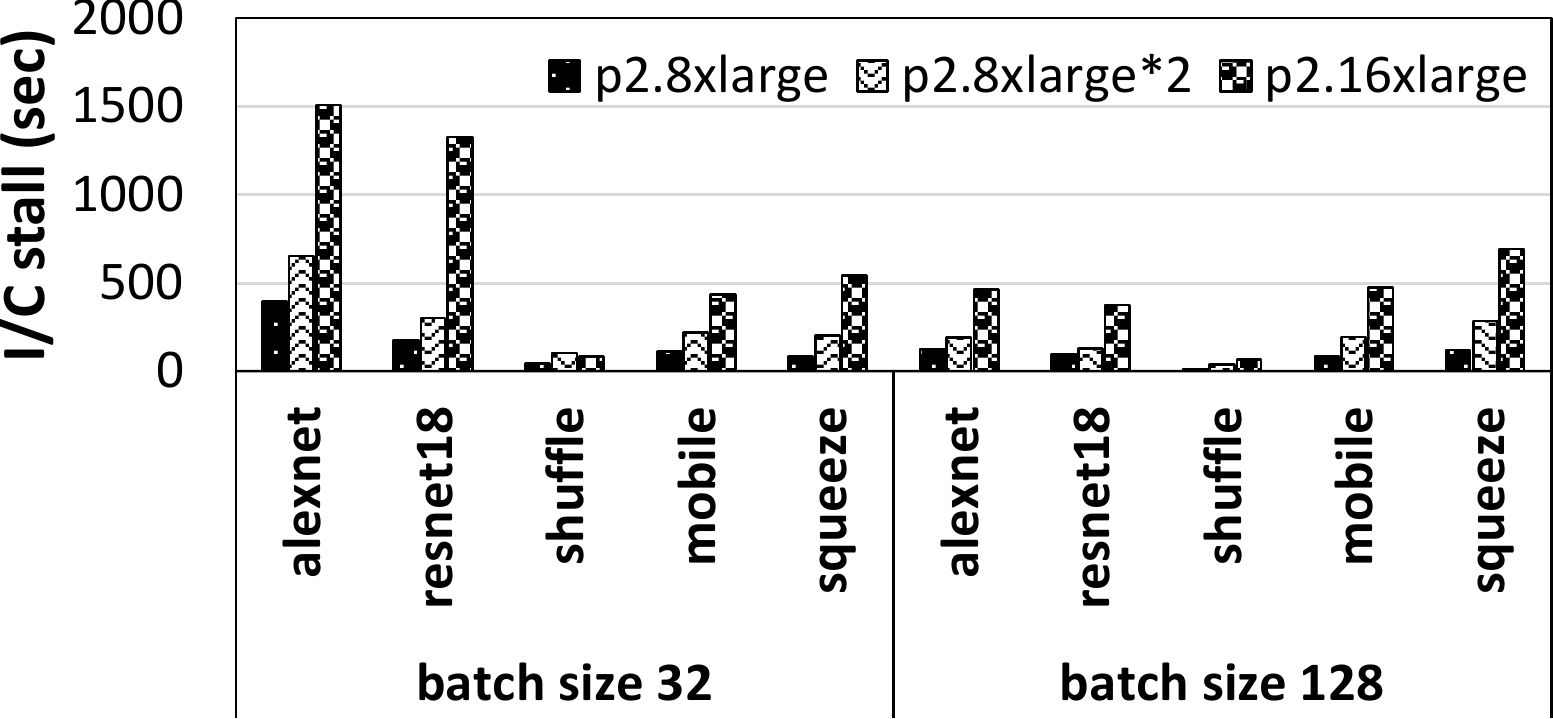} }}
    \subfloat[Interconnect stall \%]{{\includegraphics[width=0.5\linewidth]{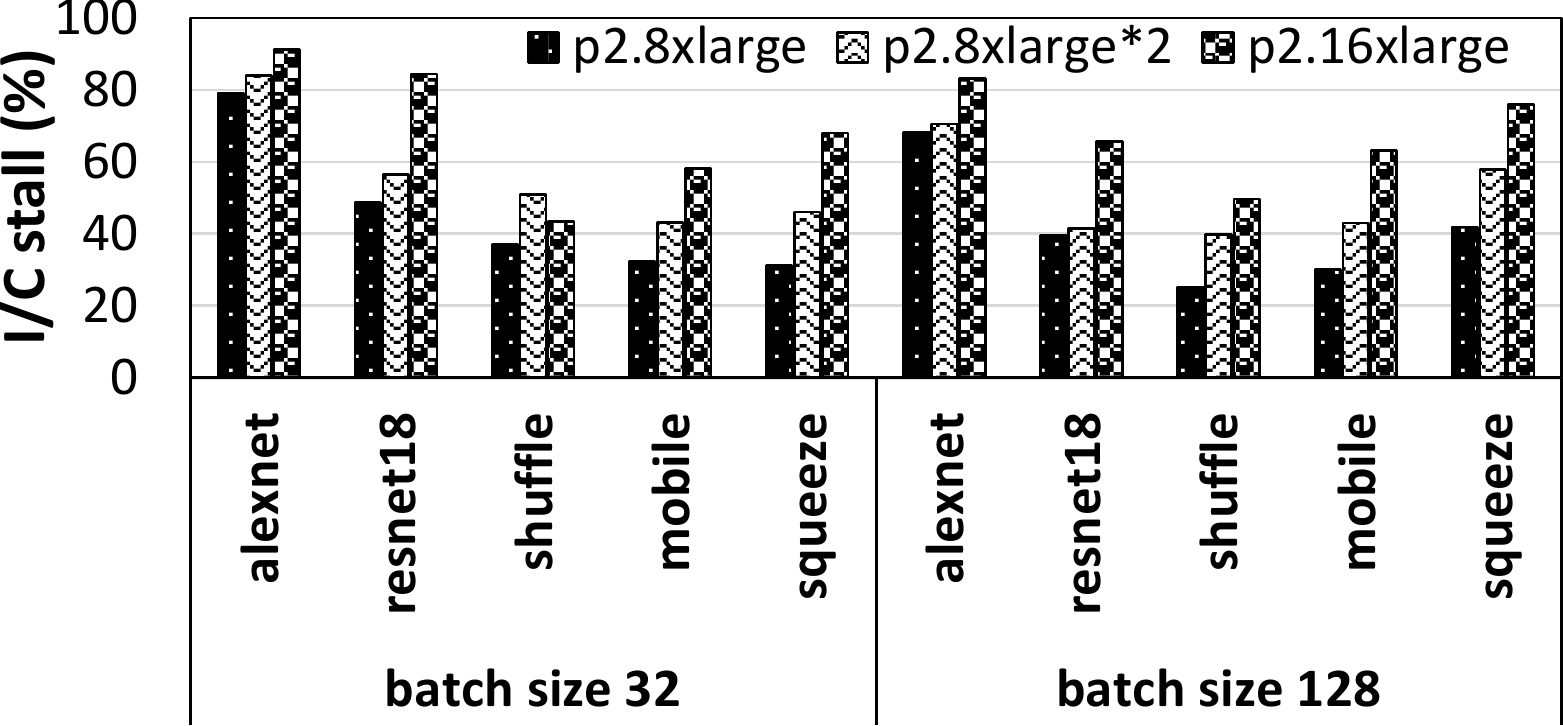} }}
    \vspace*{-3mm}
    \caption{Interconnect Stall for P2, Small Models (16xlarge has the worst stalls due to PCIe contention.)}
    \label{fig:p2_interconnect_stall}
\end{figure*}

\begin{figure}
\centering
    \subfloat[Training time (sec)]{{\includegraphics[width=0.5\linewidth]{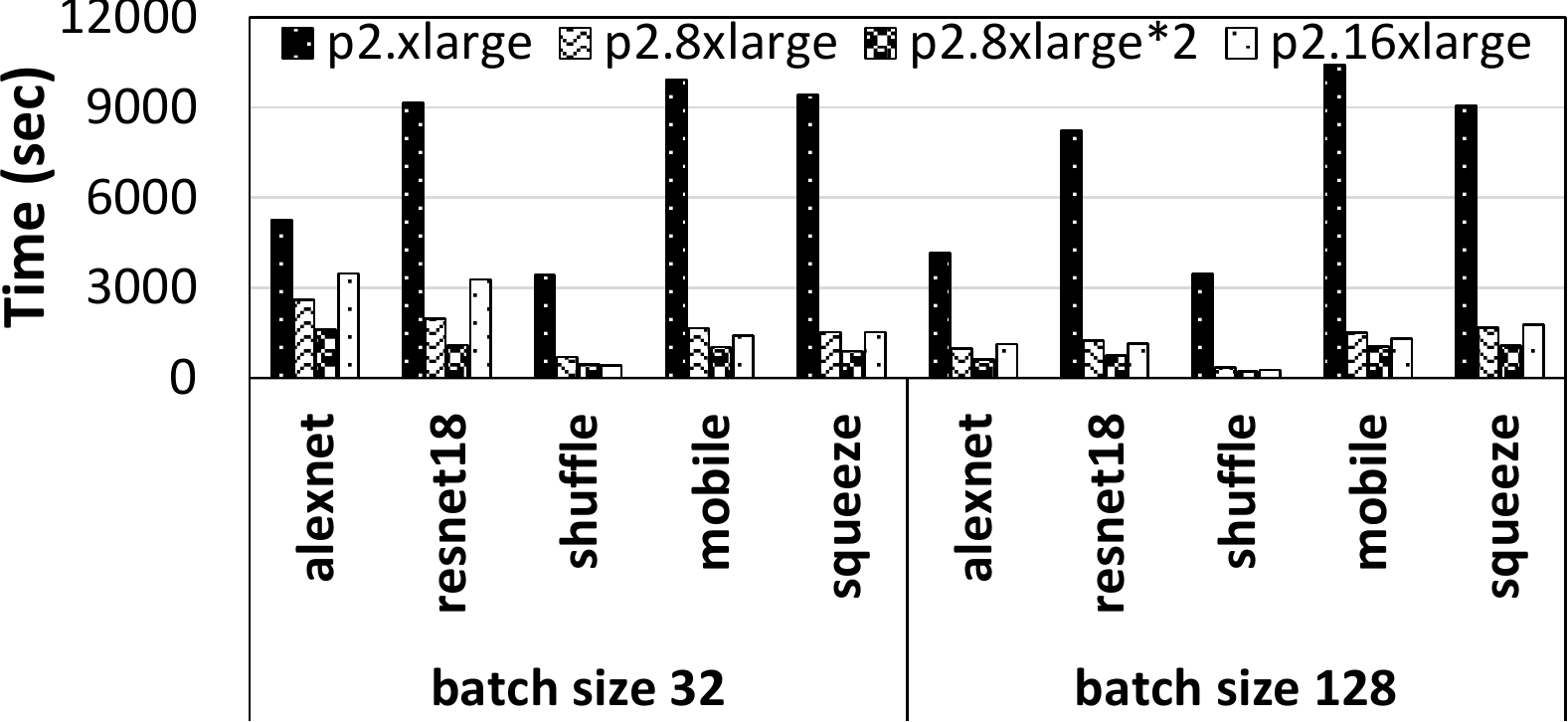} }}
    \subfloat[Training cost (\$)]{{\includegraphics[width=0.5\linewidth]{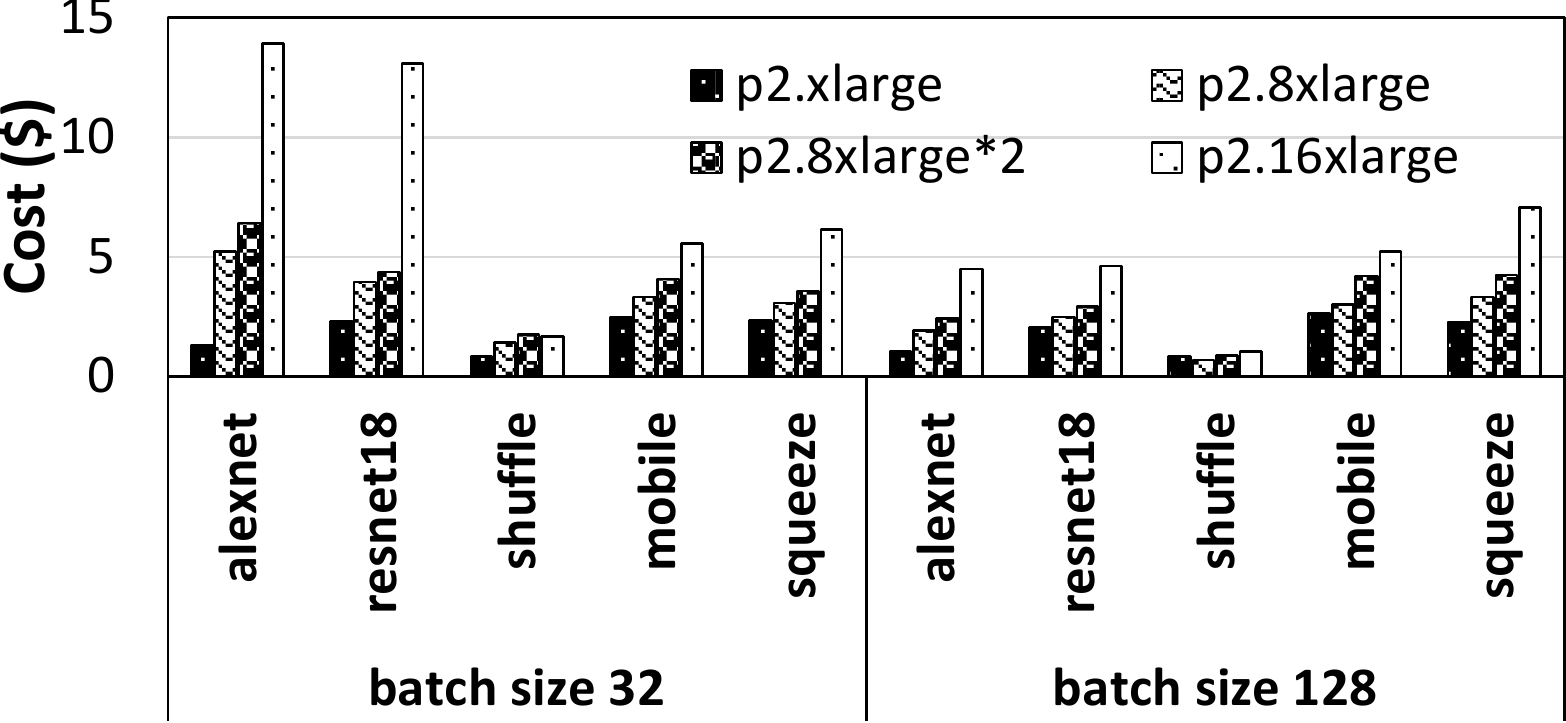} }}
    \vspace{-3mm}
    \caption{Training Time and Cost for P2, Small Models. (16xlarge is the least cost-optimal)}
    \label{fig:p2_train_time_cost_small}
    \vspace{-4mm}
\end{figure}

\vspace{-2mm}
\subsection{Analysis on AWS P2} \label{subsections: p2_evaluation}
AWS P2 instances use the NVIDIA K80 GPU with PCIe third generation interconnects.
The P2 instances consist of three instance types -- p2.xlarge, p2.8xlarge and p2.16xlarge as discussed in Section~\ref{sections: background}.
We profile P2 instances with models AlexNet, ResNet18, ShuffleNet\_v2, MobileNet\_v2 and SqueezeNet across four mini-batch sizes -- 32, 64, 96 and 128.
Since K80 GPUs have limited compute and memory resources, they are not very suitable for running large models, i.e. models with a high parameter count.
In practice, we observed very high I/C stall and monetary cost of training large models on P2. 
For e.g., for ResNet50, interconnect stall was observed to be 750\% and monetary cost was \$41 to train for a single epoch (the latter being 2000\% more than P3).
As a result, we employ the smaller models to characterize stalls on the K80 GPUs.

\subsubsection{\textbf{Stall Analysis}}
Figure \ref{fig:p2_cpu_disk_stall} shows the CPU and disk stalls as a percentage of the total training time for mini-batch sizes 32 and 128.
Unlike \cite{mohan2021}, we notice negligible CPU stalls in AWS, pointing to the fact that vCPUs at AWS are sufficient for most pre-processing needs of DL jobs.
We further notice the largest amount of disk stalls for the 16x type machine.
This is because there are 16 data loading workers running on the 16x machine to exploit the 16 GPUs of the machine.
The 16 workers read from the attached SSD in parallel and create an I/O contention.

\begin{figure}
    \centering
    \includegraphics[width=0.3\textwidth]{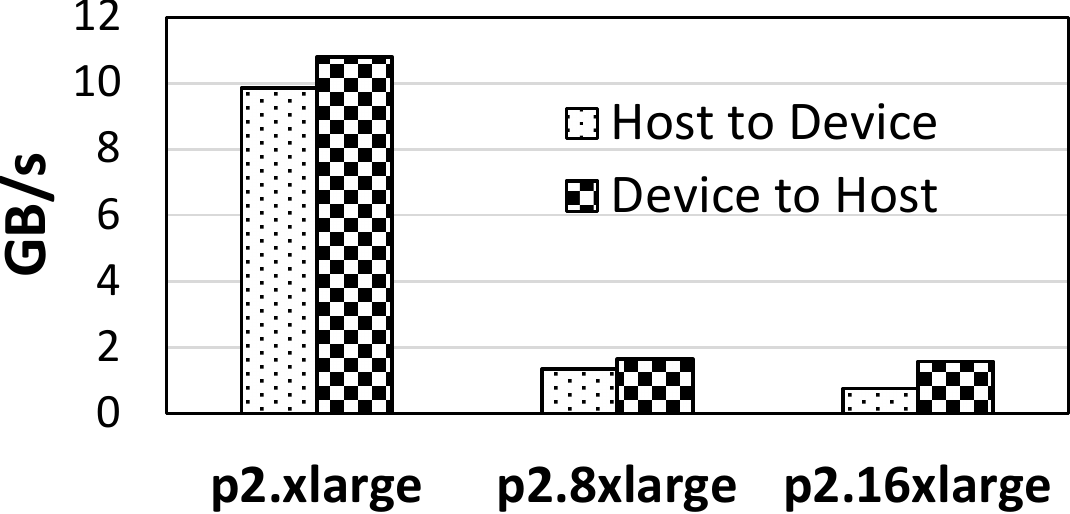}
        \caption{Per GPU PCIe bandwidth measured in P2}
\label{fig:p2_pcie}
\end{figure}

The AWS general purpose SSD used in our experiments is unable to keep up with this demand and the training spends a significant amount of time performing disk I/O (only when data is not cached in DRAM).
\par
  
We now discuss the interconnect and network stalls of P2 instances.
We observe from Figure \ref{fig:p2_train_time_cost_small}(a) that the 16x large runs a slower training than two 8xlarge machines 
that are network-connected.
This is true in all our batch runs.
Furthermore, we observe from Figure \ref{fig:p2_interconnect_stall} that 16xlarge has a higher interconnect stall time than both 8xlarge and 8xlarge*2 (two 8xlarge).
Two 8xlarge instances connected via the network are seen to not suffer from any network stalls as they are faster than the 16xlarge.
This begs the question, \textit{what is causing the slowdown in the 16xlarge?}\par

The slowdown in 16xlarge can be attributed to the limited bandwidth of the PCIe buses of P2 instances used for communication.
In case of the 16xlarge, the PCIe bandwidth is shared among 16 workers causing congestion and "slicing" of the limited PCIe bandwidth.
We validate this claim by measuring the PCIe bandwidth available per GPU using CUDA in xlarge, 8xlarge and 16xlarge instances.
All GPUs are used in parallel when running the bandwidth test and we report the per device bandwidth in Figure \ref{fig:p2_pcie}.
As is clear from the figure, the GPUs in 16xlarge instance receive significantly less bandwidth than the GPUs of all other P2 instance types. This bandwidth is  lower than the expected network bandwidth and hence the training gets throttled on the interconnect link, rather than on the network.
As the network is not the slowest link and the 8xlarge instance has access to higher interconnect bandwidth than the 16xlarge, the 8xlarge*2 configuration performs better than the 16xlarge.
This gives us an intuition that the 16xlarge instance is the least cost-optimal and we test this empirically by observing the monetary cost of running the workloads.
\par

We show the dollar cost comparison of all P2 instances in Figure \ref{fig:p2_train_time_cost_small}(b).
A linear increase in cost  is observed as  the size of the P2 instance is increased.
This confirms the intuition from our study of interconnect stalls that the monetary
cost of executing a DDL workload is proportional to the observed interconnect stall of that workload.
The lowest cost of running the training is on P2.xlarge, which has a single GPU and hence, has no interconnect stalls.
However, the DDL execution time does not always \emph{decrease} linearly from smaller instance to larger instance.
From Figure \ref{fig:p2_train_time_cost_small}(a), we notice that there is no significant improvement in training time on 16xlarge for a 2$\times$ increase in cost.
In fact, we notice in most cases that the running time in 16xlarge is more than that of 8xlarge, although the instance has twice the resources as that of 8xlarge.
This is because although resources like CPU, GPU and memory are doubled, the PCIe bus bandwidth remains the same (as already demonstrated), thereby causing congestion and significant slowdowns.

\subsubsection{\textbf{Recommendation}}
We observe both high interconnect and disk stalls on the 16xlarge instance and accordingly, believe the 16xlarge instance may not be the cost-optimal choice.
Even when more GPUs are needed than what the 8xlarge instance can provide, training time and cost are lower when using a combination of 8xlarge instances connected via network compared to using the 16xlarge instance.

\subsection{Analysis on AWS P3} \label{subsections: p3_eval}
AWS P3 instances use the NVIDIA V100 GPU with NVLink interconnect as already described in Section~\ref{sections: background}.
The P3 instances are high-performing instances capable of training large DNN models in a cost-effective manner.
We begin our discussion with the stall analysis of P3 in the sequel.

\begin{figure}
\centering
    \subfloat[CPU stall \% (CPU stall is negligible)]{{\includegraphics[width=0.5\linewidth]{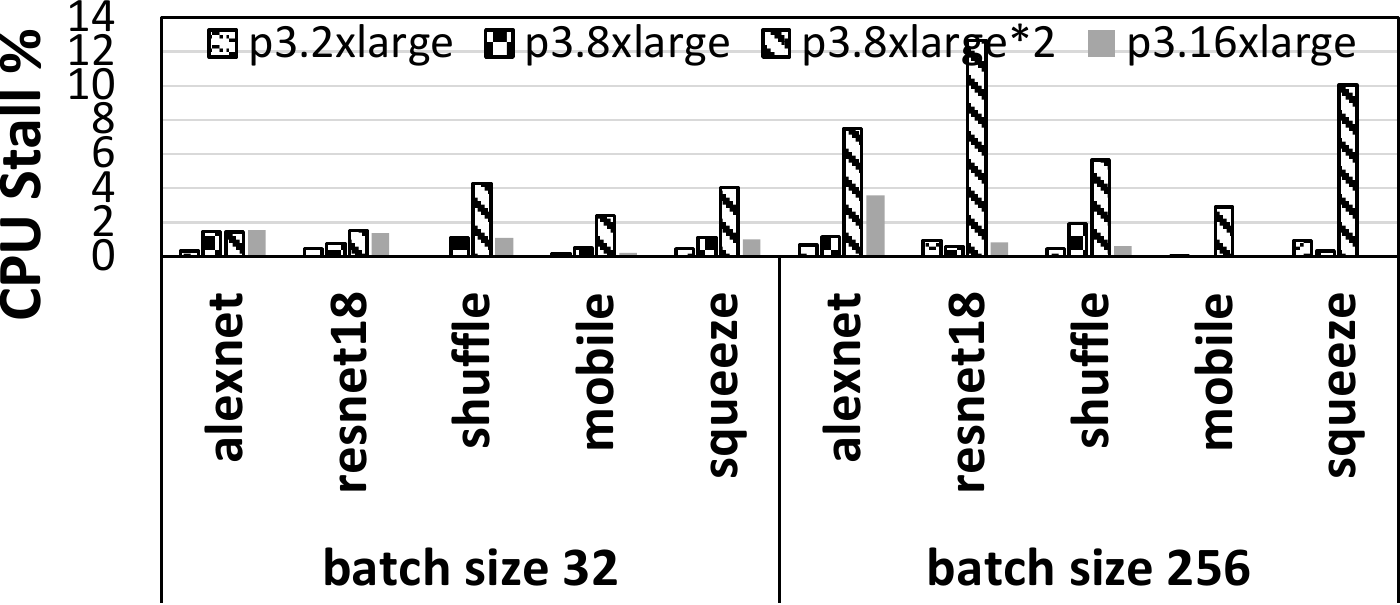} }}
    \subfloat[Disk stall \% (Disk stall highest for 16xlarge)]{{\includegraphics[width=0.5\linewidth]{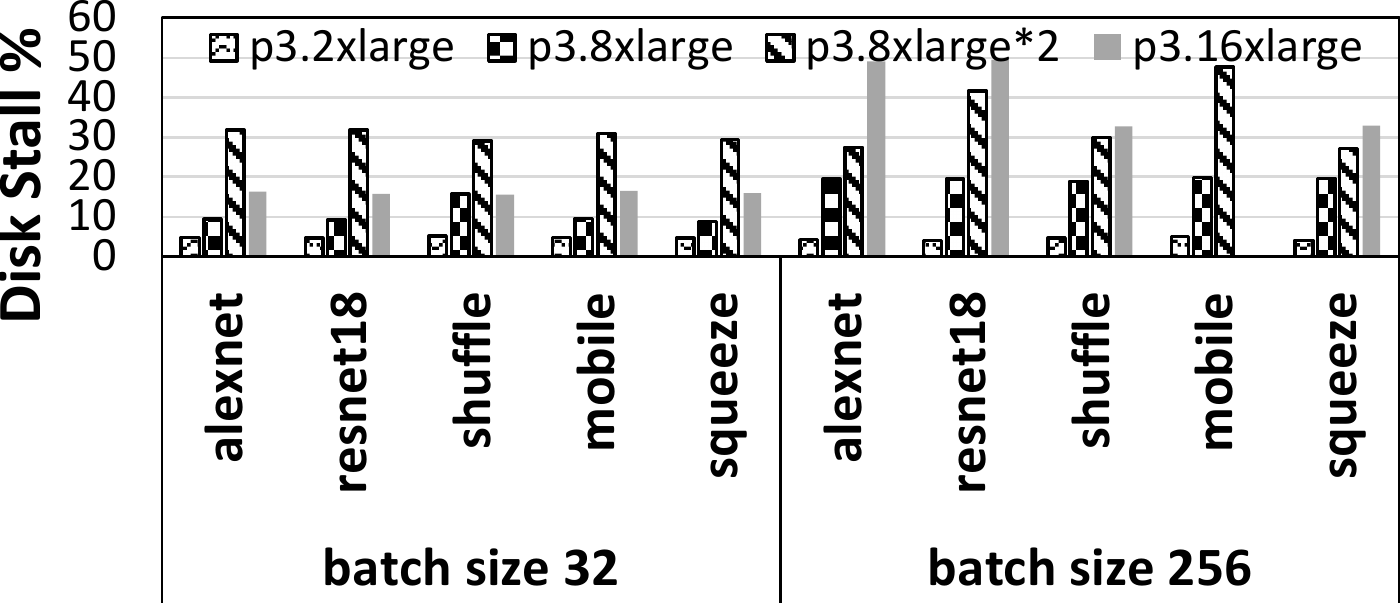} }}
    \vspace{-3mm}
    \caption{CPU and disk stall for P3, small models.}
    \label{fig:p3_cpu_disk_stall_small}
    \vspace{-4mm}
\end{figure}

\begin{figure}
\centering
    \subfloat[CPU stall \%]{{\includegraphics[width=0.49\linewidth]{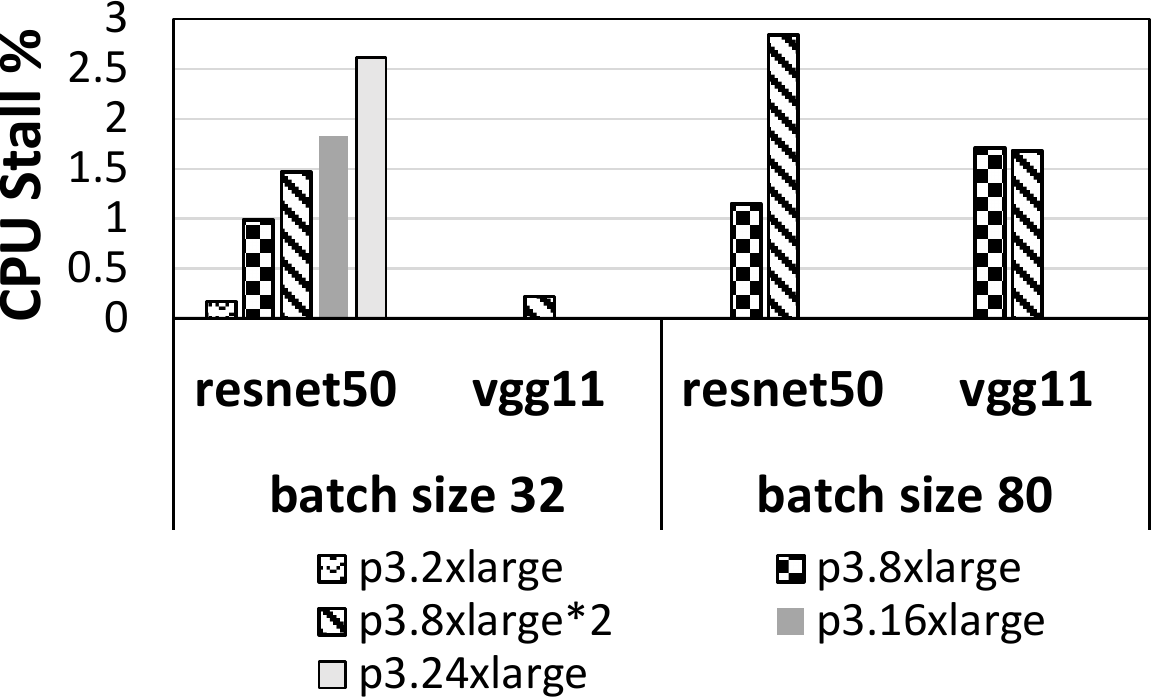} }}
    \subfloat[Disk stall \%]{{\includegraphics[width=0.49\linewidth]{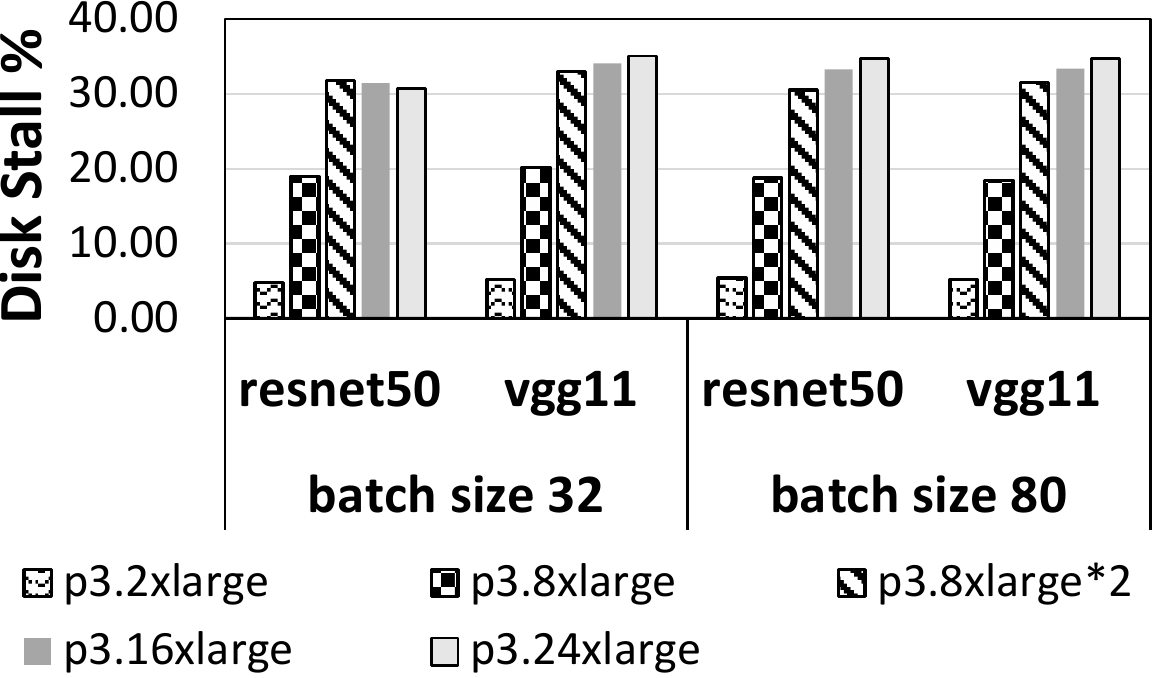} }}
    \vspace{-3mm}
    \caption{CPU and disk stall for P3, Large image models. (CPU stall is negligible, disk stall high for experiments with 8 GPUs)}
    \label{fig:p3_cpu_disk_stall_large}
    \vspace{-4mm}
\end{figure}

\begin{figure}
\centering
    \subfloat[Interconnect stall (sec)]{{\includegraphics[width=0.49\linewidth]{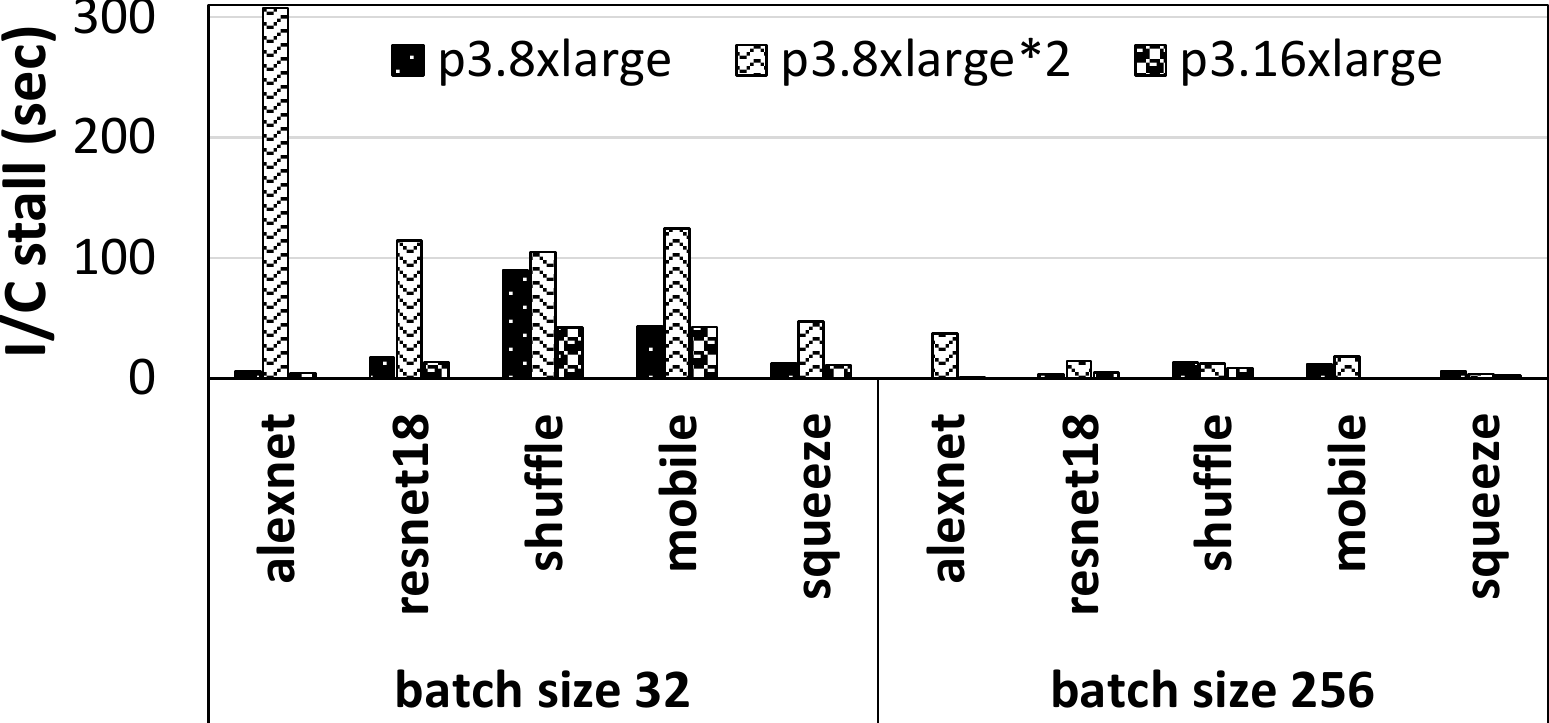} }}
    \subfloat[Interconnect stall \%]{{\includegraphics[width=0.49\linewidth]{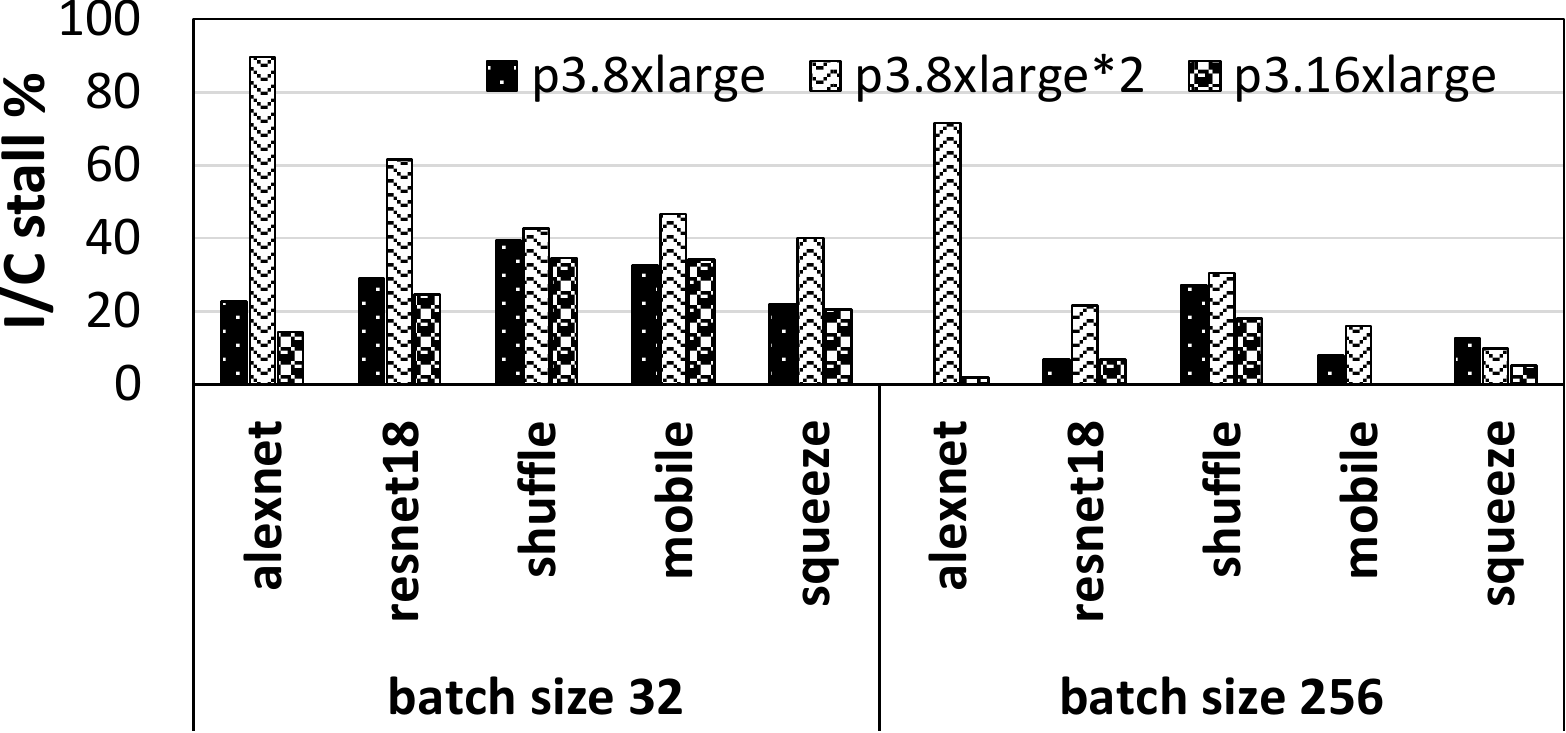} }}
    \vspace{-3mm}
    \caption{Interconnect stall for P3, Small models. (8xlarge suffers from sub-optimal interconnect allocation)}
    \label{fig:p3_interconnect_stall_small}
    \vspace{-3mm}
\end{figure}

\begin{figure}
\centering
    \subfloat[Training time (sec)]{{\includegraphics[width=0.49\linewidth]{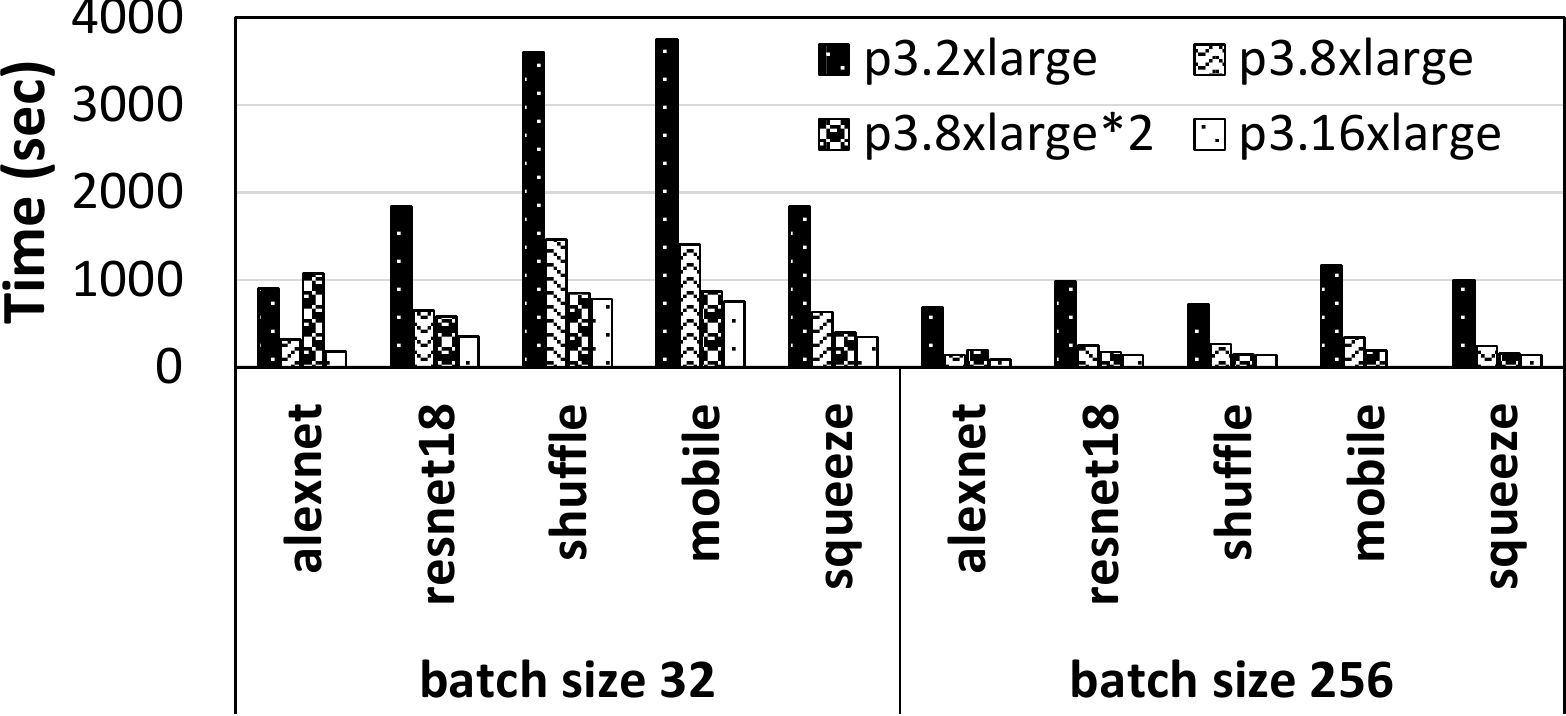} }}
    \subfloat[Training cost (\$)]{{\includegraphics[width=0.49\linewidth]{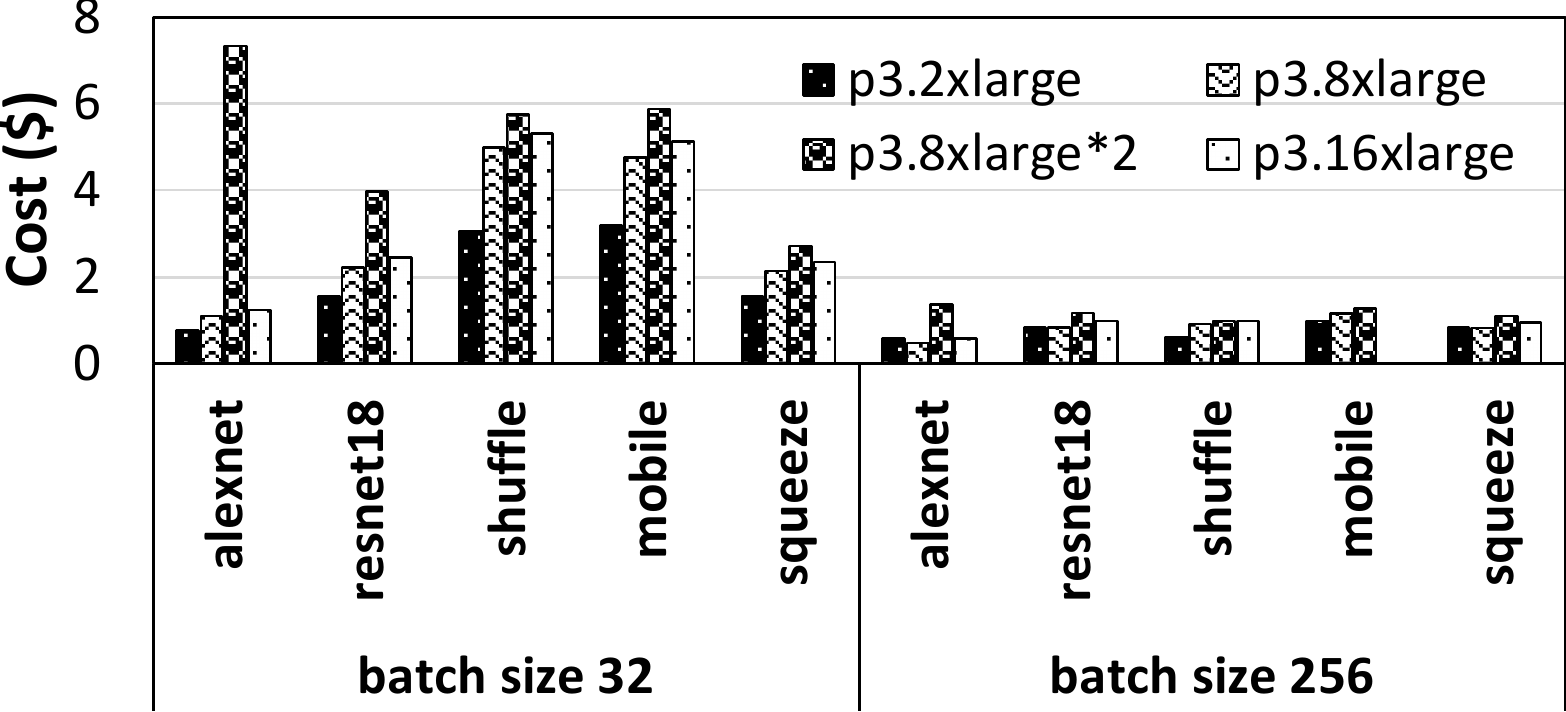} }}
    \vspace{-3mm}
    \caption{Training Time and Cost for P3, Small Models. (16xlarge is the most performant)}
    \label{fig:p3_train_time_cost_small}
    \vspace{-3mm}
\end{figure}

\begin{figure}
\centering
    \subfloat[Batch size 32]{{\includegraphics[width=0.49\linewidth]{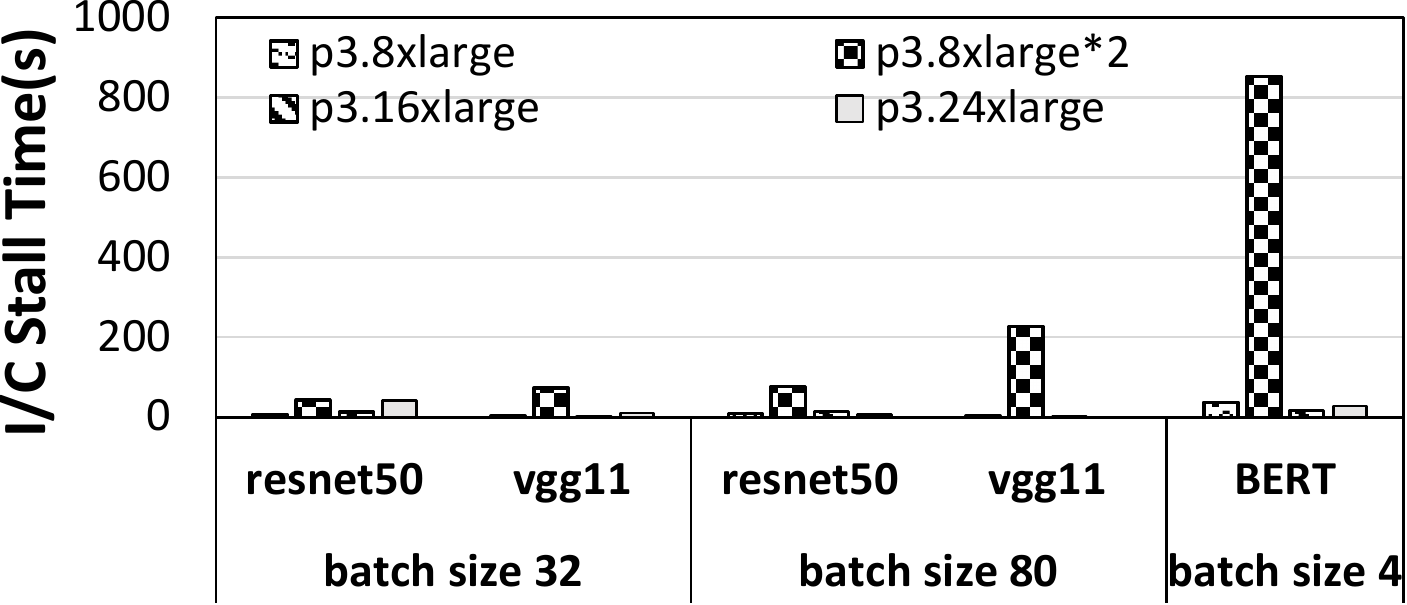} }}
    \subfloat[Batch size 80]{{\includegraphics[width=0.49\linewidth]{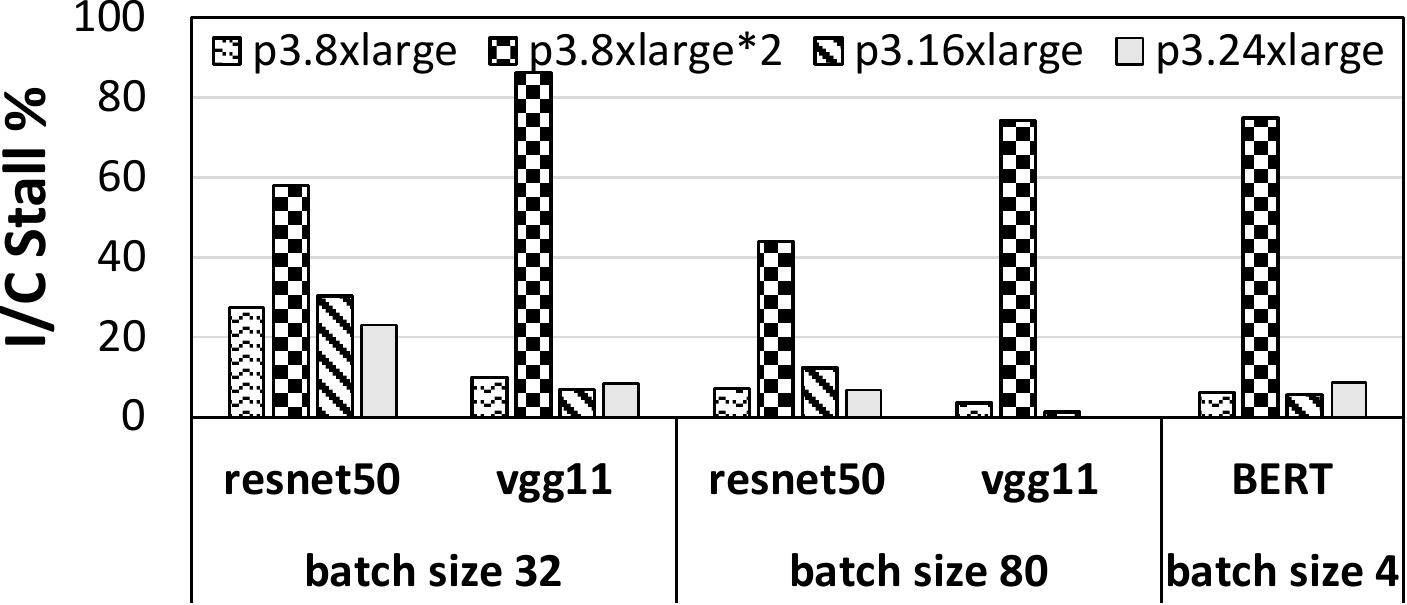} }}
    \vspace{-3mm}
    \caption{Interconnect stall for P3, Large Models. (16xlarge has the lowest stall)}
    \label{fig:p3_interconnect_stall_large}
    \vspace{-6mm}
\end{figure}

\begin{figure}
\centering
    \subfloat[Training time (sec)]{{\includegraphics[width=0.49\linewidth]{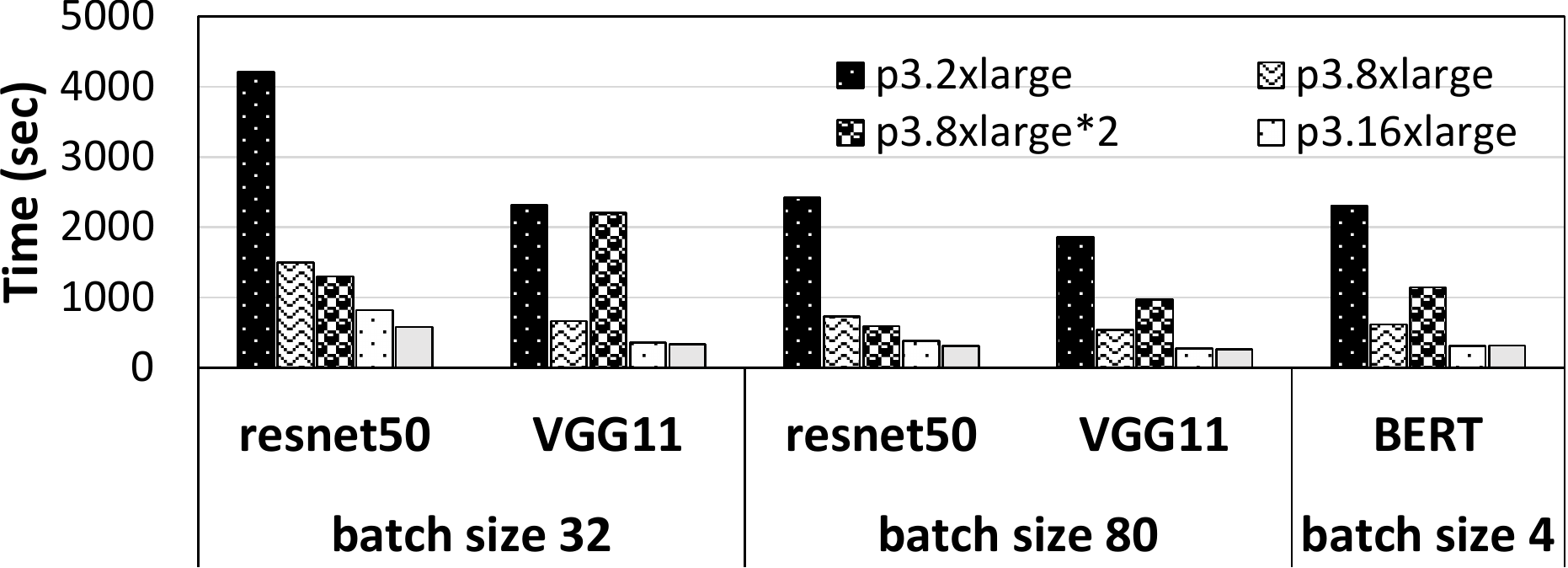} }}
    \subfloat[Training cost (\$)]{{\includegraphics[width=0.49\linewidth]{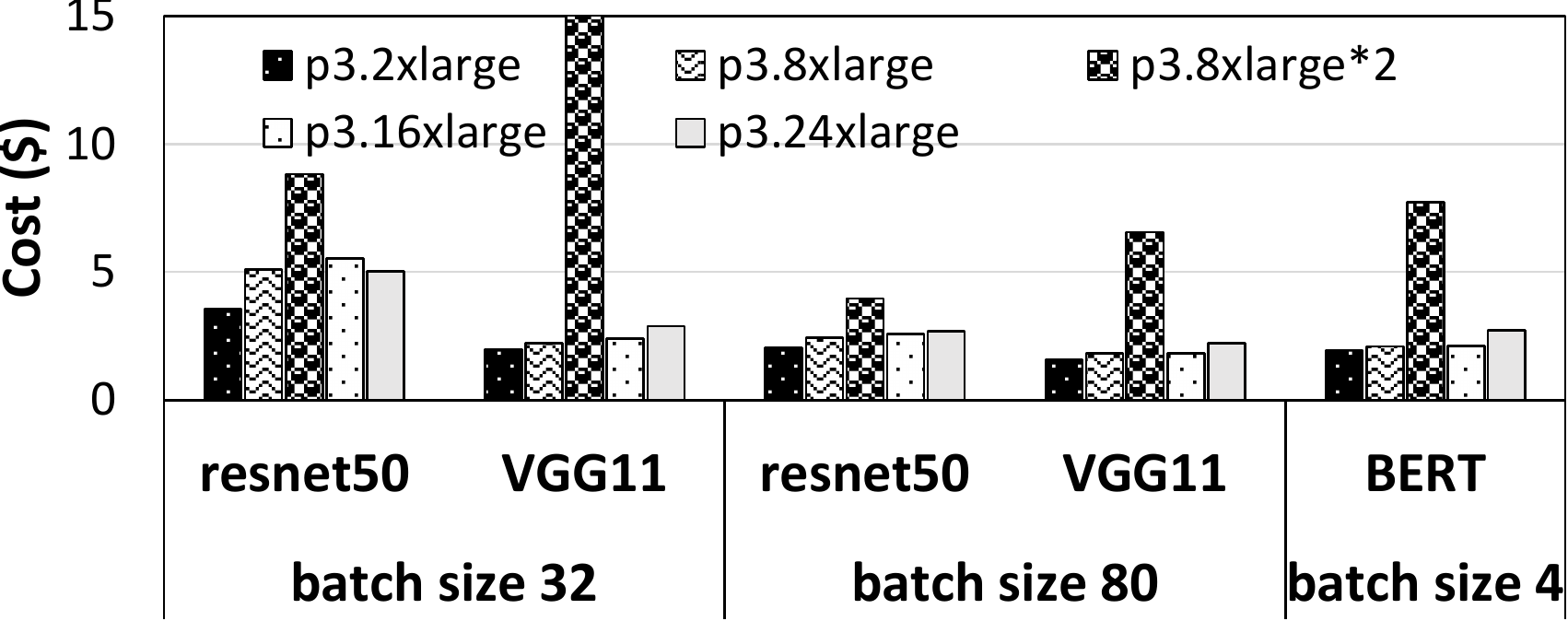} }}
    \vspace{-3mm}
    \caption{Training Time and Cost for P3, Large Models. (16xlarge and 24xlarge are equally performant)}
    \label{fig:p3_train_time_cost_large}
    \vspace{-4mm}
\end{figure}

\subsubsection{\textbf{Stall Analysis}}
We show CPU and disk stalls for small models in Figure \ref{fig:p3_cpu_disk_stall_small} and large models in Figure \ref{fig:p3_cpu_disk_stall_large}.
The CPU and disk stalls follow the same pattern as in P2.
The CPU stalls are negligible and the disk stall is high for the 16xlarge instance.
Unlike p2.16xlarge, the p3.16xlarge has eight GPUs and hence, eight workers perform I/O on the attached SSD.
However, the throughput of training is also high due to the higher compute capacity (of both GPU and CPU) of the instance, thus, leading to higher usage of the SSD and higher disk stalls (data is not cached in DRAM).\par

The P3 instances use NVLink for communication between the GPUs instead of the PCIe bus.
As discussed in Section~\ref{sections: background}, NVLink offers significantly higher bandwidth compared to traditional PCIe-based communication and hence, we expect lower interconnect stalls while using NVLink.
We measure and show the actual interconnect stalls for P3 in Figure \ref{fig:p3_interconnect_stall_small} and \ref{fig:p3_interconnect_stall_large} and notice that they are lower than those of the P2 instances, as expected.
However, we also observe the 8xlarge (which has half the number of GPUs as the 16xlarge) to have higher overall interconnect stalls than the 16xlarge, especially for smaller models or while using smaller batch sizes. 
As the number of GPUs decreases, the volume of gradients to be transferred (as each GPU generates gradients) also decreases, thereby, requiring lesser bandwidth from the underlying interconnect. This should ideally translate into lower interconnect stalls for the 8xlarge.
Therefore, we ask the question: \textit{why does the p3.8xlarge instance not have strictly lower interconnect stalls than the 16xlarge?}

The reason for this anomaly is that although AWS provides a highly connected crossbar architecture (refer Figure \ref{fig:p3_interconnect}) for communication via the NVLink, this may not be the case for the 8xlarge.
Ideally, AWS should split the 16xlarge instance into two 8xlarge instances such that each instance gets an entire crossbar as shown by the dotted line in Figure \ref{fig:p3_interconnect}.
This would have provided the tenant/user with a highly-connected, high bandwidth GPU interconnect, resulting in lower interconnect stalls.
However, we theorize that AWS is not able to "evenly slice" the physical interconnect so as to give an entire crossbar to the 8xlarge instance.
This may be due to multiple single size GPU requests from several tenants occupying GPUs in a crossbar.
The 8xlarge instance loses the benefit of the crossbar architecture due to this and ends up being less performant with respect to interconnect stalls.
This "trait" of AWS interconnects is essentially probabilistic in nature and a tenant may indeed end up getting an entire crossbar for their 8xlarge instance, thereby, resulting in lower interconnect stalls.\par

Next, we compare the performance of p3.16xlarge with that of the p3.24xlarge.
The p3.24xlarge is a dedicated instance offering which has the same number and type of GPUs as the 16xlarge but with twice the memory. 
It also comes with a dedicated local SSD storage along with more vCPUs and DRAM than the 16xlarge (refer Table \ref{tab: public cloud gpu offerings}).
However, from our stall analysis of the 24xlarge, we do not observe any significant decrease in stalls or training time compared to the 16xlarge.
This is true even for our BERT large model which is both compute and memory--intensive.
We now ask: \textit{why is the performance of 24xlarge not strictly better than the 16xlarge?} \\
The answer to this question, again, lies in its GPU interconnect.
From \cite{p3_interconnect} we know that both the 16xlarge and the 24xlarge use the same NVlink interconnect hardware and hence they also suffer from the same types of interconnect stalls.
Although the 24xlarge offers a better configuration for each of its hardware components (GPU, DRAM, CPU, SSD etc.), it misses out on improving its NVLink interconnect.
The DNN pipeline suffers from the same amount of communication overhead as the 16xlarge and hence, is not able to exploit the better hardware.
This further lends credence to the importance of communication overhead in DDL (missed by prior work).\par

\begin{figure}
\centering
    \subfloat{{\includegraphics[width=0.4\textwidth]{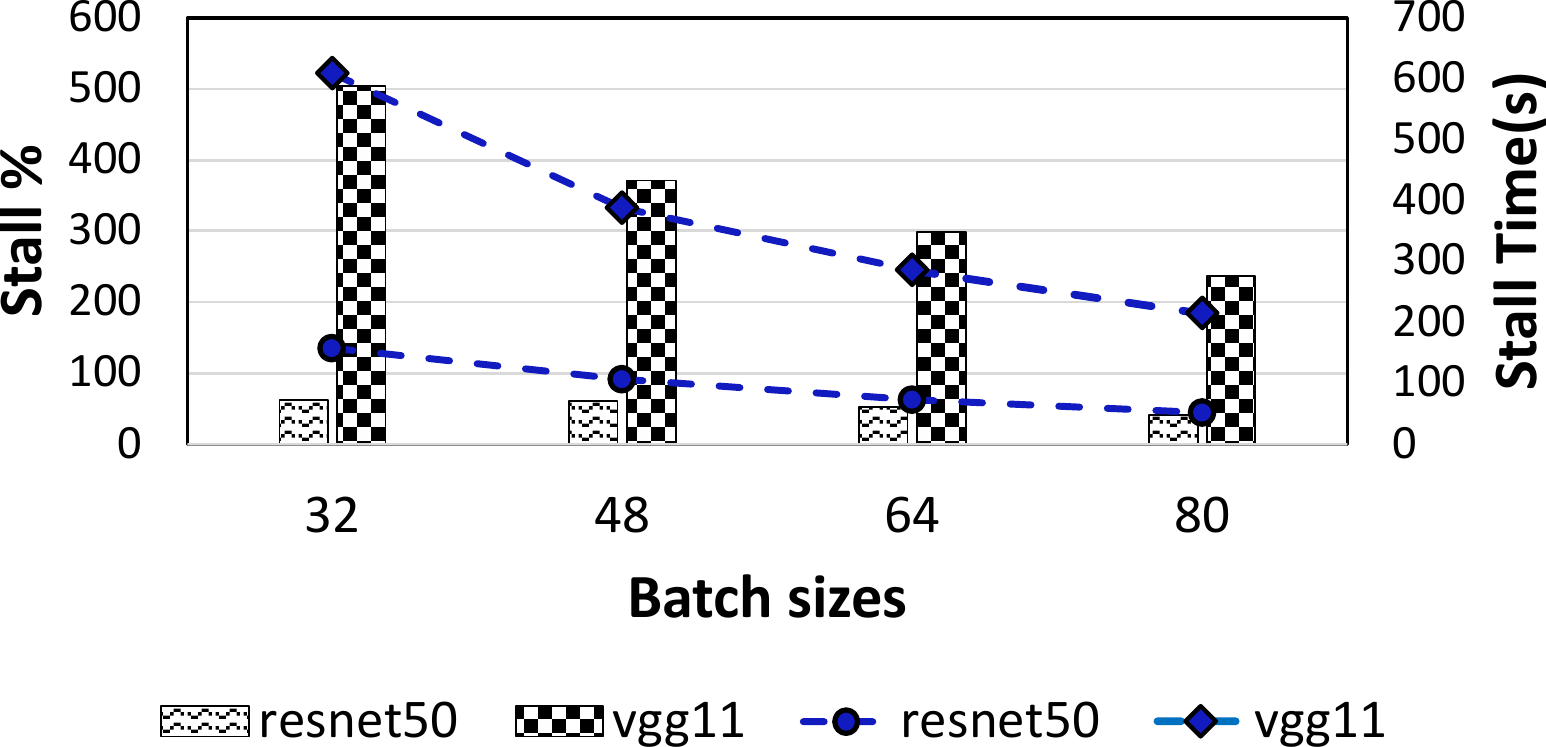} }}
\caption{Network stall of two p3.8xlarge instances. (Network stall is as high as 500\%)}
\label{fig:network_overhead}
\end{figure}

However, there is a caveat to this. The 24xlarge instance has twice the amount of per-GPU memory (32GB) than the 16xlarge.
This allows users to run training with larger batch sizes thereby reducing time per epoch.
However, we can't conduct a cost analysis between 16xlarge and 24xlarge by increasing the batch size of training on 24xlarge due to two reasons: (i) single epoch with different batch sizes is not representative of the same end-to-end training, and (ii) large batch sizes tend to converge to sharp minimizers which leads to poor generalization \cite{keskar2016large}.
But for comprehensiveness, we run our BERT model on the 24xlarge after doubling the batch size to 8. This resulted in about 12.8\% improvement in training time and costing about \$2.37. 
This is still more than the \$2.1 cost of running the model on 16xlarge with half the batch size. \\
Finally, we ask: \textit{what happens when the instances are connected via the network?}
To answer this question, we calculate the network stall of two p3.8xlarge instances connected via the network (p3.8xlarge*2) in Figure \ref{fig:network_overhead} as part of step~\tcircled{5} of \name and notice network stalls as high as 500\%.
This is because as soon as the "all-reduce" ring contains a network link, the training gets throttled on this slow network link.
Compared to the NVLink interconnect, which has a sufficiently large bandwidth to accommodate fast data transfers, the network link introduces higher slowdowns.
This discourages us to run training over network links. \par

Note that we do observe large models like VGG to have low interconnect stall (but high network stall). The reasoning for this will be discussed in Section~\ref{subsections: micro_characterization}.

\subsubsection{\textbf{Cost Analysis}}
We show the cost and time analysis of P3 instances in Figures \ref{fig:p3_train_time_cost_small} and \ref{fig:p3_train_time_cost_large}.
The cost analysis follows the same pattern as that in P2 instances but the performance of the instances differs.
We find that the smallest P3 instance, the 2xlarge is the most cost optimal followed by the 8xlarge and the 16xlarge.
The 24xlarge is the least cost-optimal in most experiments.
An immediate question that can be asked here is: \textit{how is 8xlarge more cost optimal than both 16xlarge and the 24xlarge?} \\
The answer to this question is that although 16x/24xlarge instances have lower interconnect stalls than the 8xlarge, they still suffer from higher disk stalls (due to more number of workers, refer Figures \ref{fig:p3_cpu_disk_stall_small}(b) and \ref{fig:p3_cpu_disk_stall_large}(b)) and hence, end up being less cost-effective than the 8xlarge.
Note that the actual disk stall suffered is not as high as shown in the disk stall analysis due to caching of data. 
The disk stall is only high enough to compensate for the small interconnect stall difference between 8xlarge and the 16xlarge.
It is mostly the interconnect stall that drives the cost-effectiveness of an instance.
We also observe that the network connected instances are the least cost optimal due to high network stalls.
\vspace{-1mm}
\subsubsection{\textbf{Recommendation}}
We recommend the single 2xlarge as the most cost-effective instance for training.
However, we realize that using a single GPU is not practical to train most models due to time constraints.
Hence, tenants must specifically find out the stalls for their models before running an end-to-end training on an 8xlarge or a 16xlarge.
Fortunately, \name is designed to solve this very problem and tenants can use it to find out the various stalls in their model.
We do not recommend the use of 24xlarge unless the model requires the high GPU memory offered.

\begin{figure}
\centering
    \subfloat[Training time per epoch]{{\includegraphics[width=.8\linewidth]{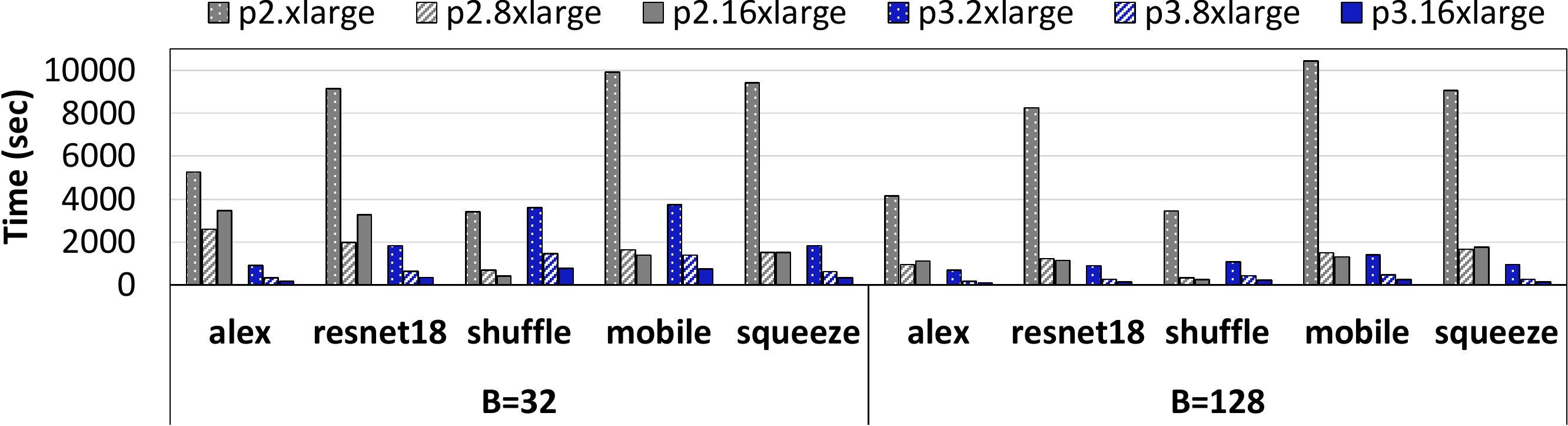} }}
    \qquad
    \subfloat[Training cost per epoch]{{\includegraphics[width=.75\linewidth]{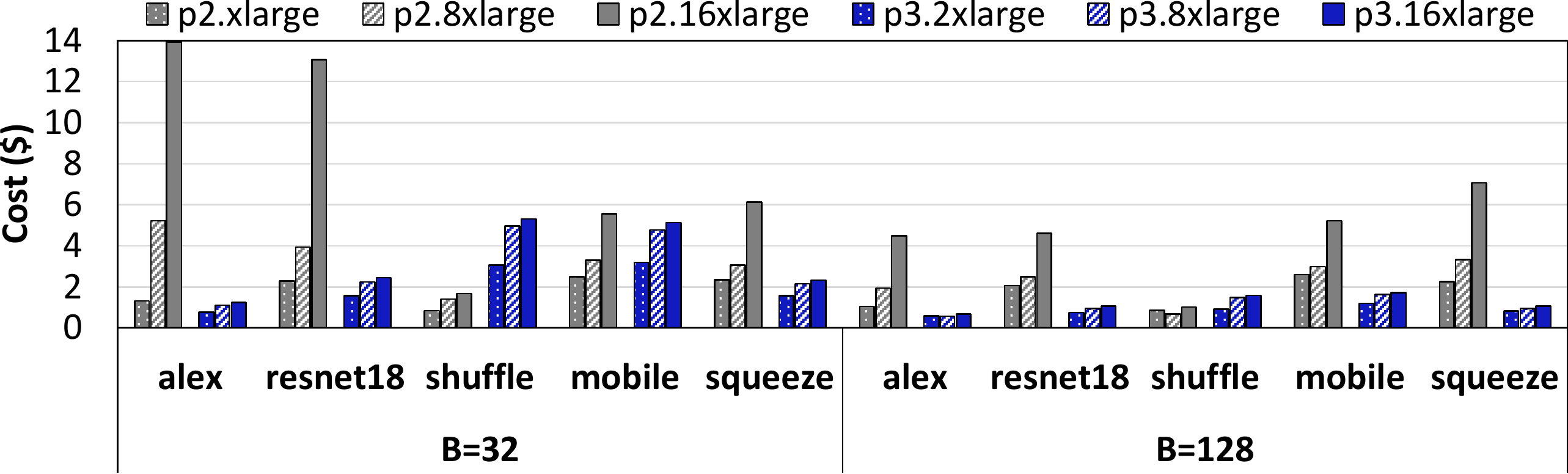} }}
    \vspace{-3mm}
    \caption{P2 vs P3 train-time/cost comparison. (P3 is generally more cost-optimal except for very small models)}
    \label{fig:p2_p3_cost_comparison}
    \vspace{-6mm}
\end{figure}

\begin{figure}
\centering
    \subfloat[GPU compute utilization]{{\includegraphics[width=0.49\linewidth]{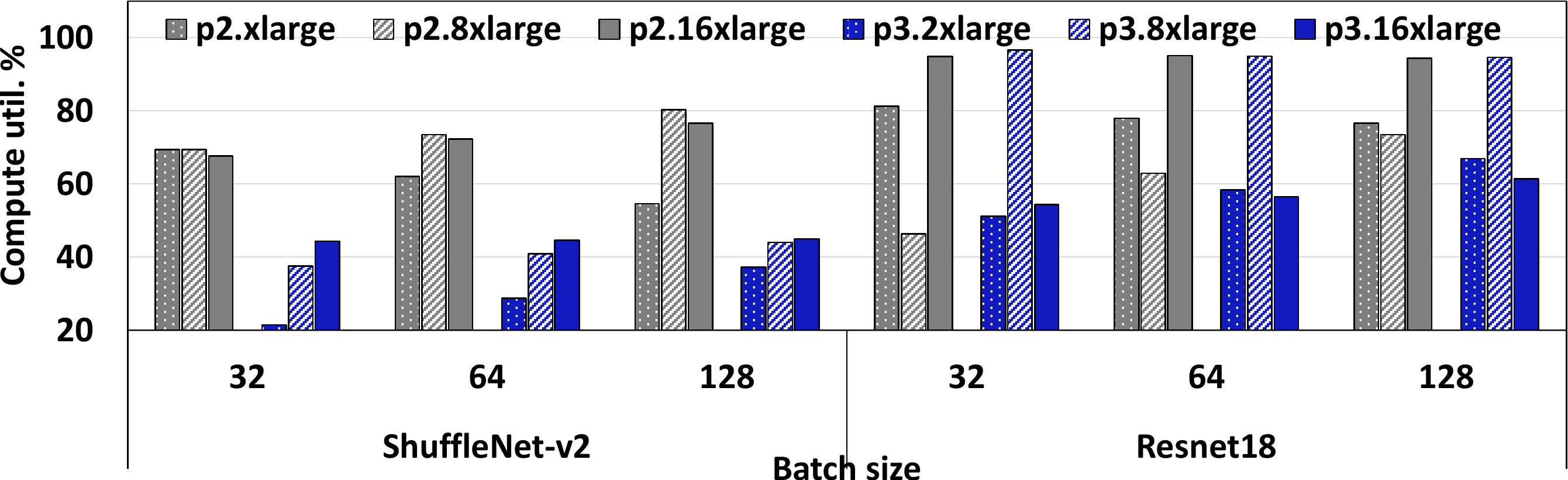} }}
    \subfloat[GPU memory utilization]{{\includegraphics[width=0.49\linewidth]{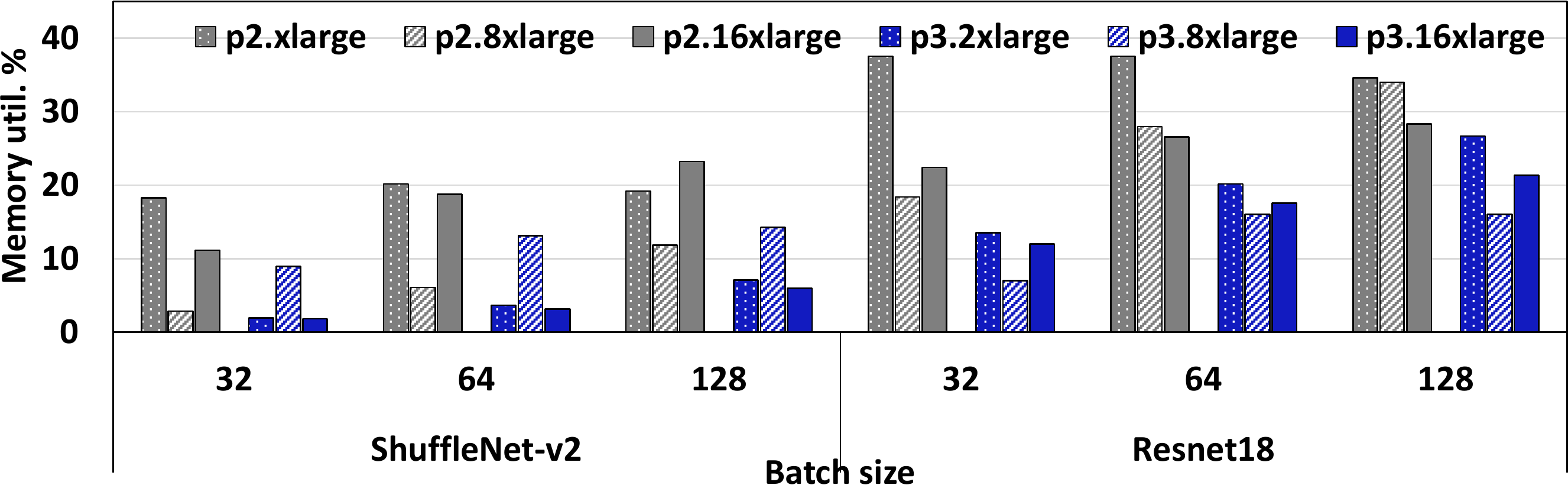} }}
    \vspace{-3mm}
    \caption{GPU util. of P2 vs P3 for a small \& large model. (Small models have low GPU util. in P3) }
    \label{fig:resnet18_vs_shufflenet}
    \vspace{-6mm}
\end{figure}

\subsection{Comparison between P2 and P3} \label{subsections: p2_vs_p3}
We now compare the two GPU instance types -- P2 and P3 from a cost-efficiency perspective.
From Figure \ref{fig:p2_p3_cost_comparison}, we notice that P3 instances are generally more cost-effective than P2 instances, although P3 instances are about 3.5$\times$ costlier per hour than P2 instances. 
This is because of the lower stalls P3 instances experience compared to their P2 counterparts.
However, some smaller models like ShuffleNet aren't able to exploit the memory and compute capacity of large V100 GPUs present in the P3 instances, unlike models with many layers like ResNet18 (shown in Figure \ref{fig:resnet18_vs_shufflenet}).
Hence, such small models incur the least cost when trained on P2 instances. Figure  \ref{fig:p2_p3_cost_comparison} shows the training time and cost of running various models on P2 and P3.

\subsubsection{\textbf{Recommendation}}
While we recommend using P3 instances whenever possible, we do notice that smaller models such 
as ShuffleNet can be trained cost-effectively on P2s.
We also note that AWS has limited GPU availability and tenants might not always receive the desired number/type of GPUs from AWS.
Thus, tenants may be forced to use P2 instances due to unavailability of P3s.

\section{Micro Characterization and Network Stall Analysis } \label{sections: evaluation_synthetic}


In this section, we analyze the interconnect and network stalls through synthetic DNN models to (i) find characteristics in model architecture that influence interconnect and network stall behavior, and (ii) express the generality of our interconnect and network stall profiler for unseen models.
We then describe network stall at scale through modelling and empirical experiments.

\begin{figure}
    \centering
    \subfloat[I/C Stall Time]{{\includegraphics[width=0.49\linewidth]{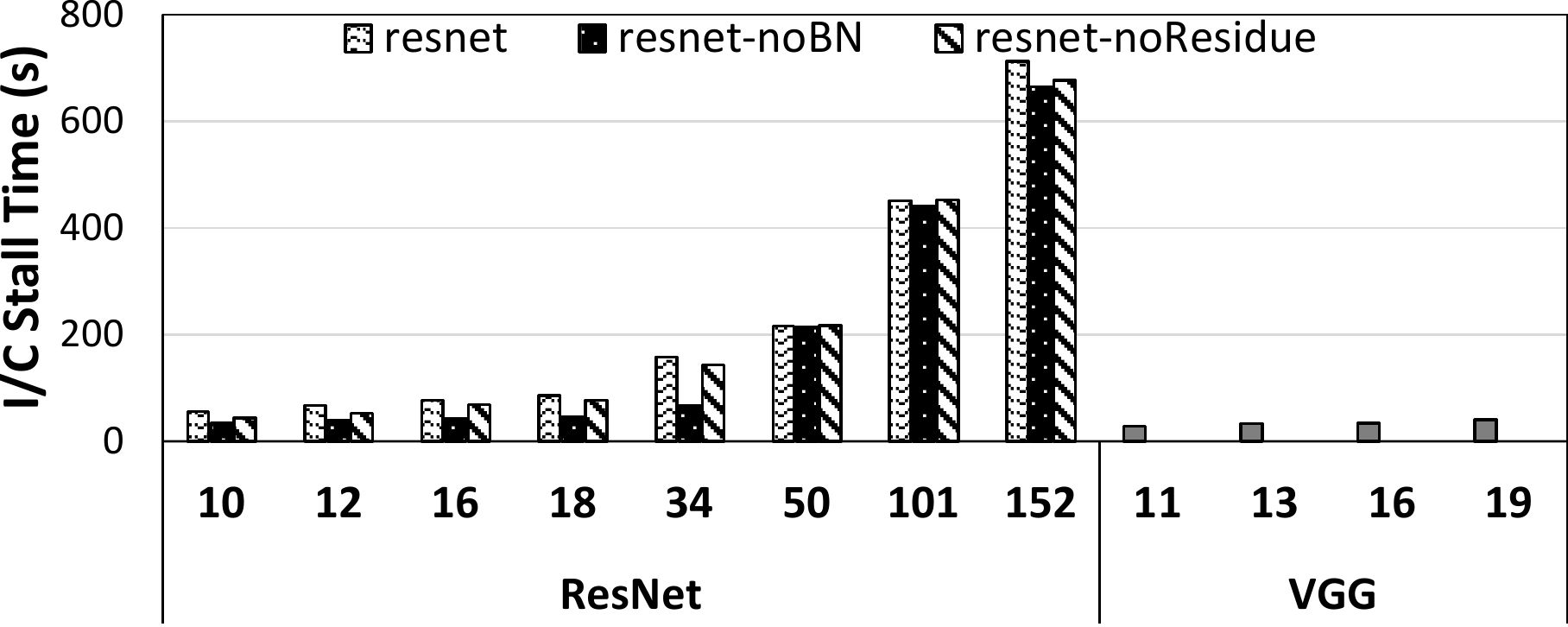} }}
    \subfloat[I/C Stall \%]{{\includegraphics[width=0.49\linewidth]{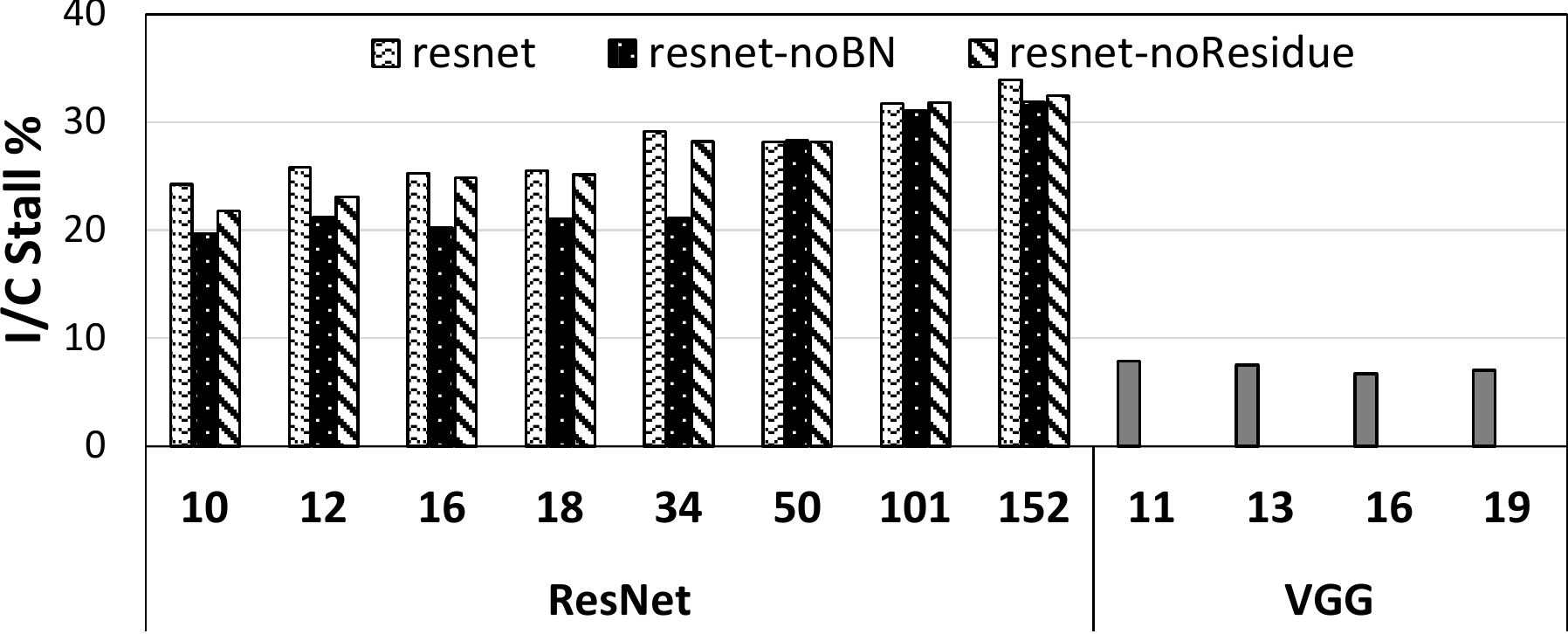} }}
    \vspace{-3mm}
    \caption{Interconnect stall in ResNet and VGG with increasing number of layers. (VGG has low interconnect stall while being much larger than ResNet)}
    \label{fig:p3_synthetic_ic_stalls}
    \vspace{-3mm}
\end{figure}


\begin{figure*}
    \centering
    \subfloat[N/W Stall Time]{{\includegraphics[width=0.49\linewidth]{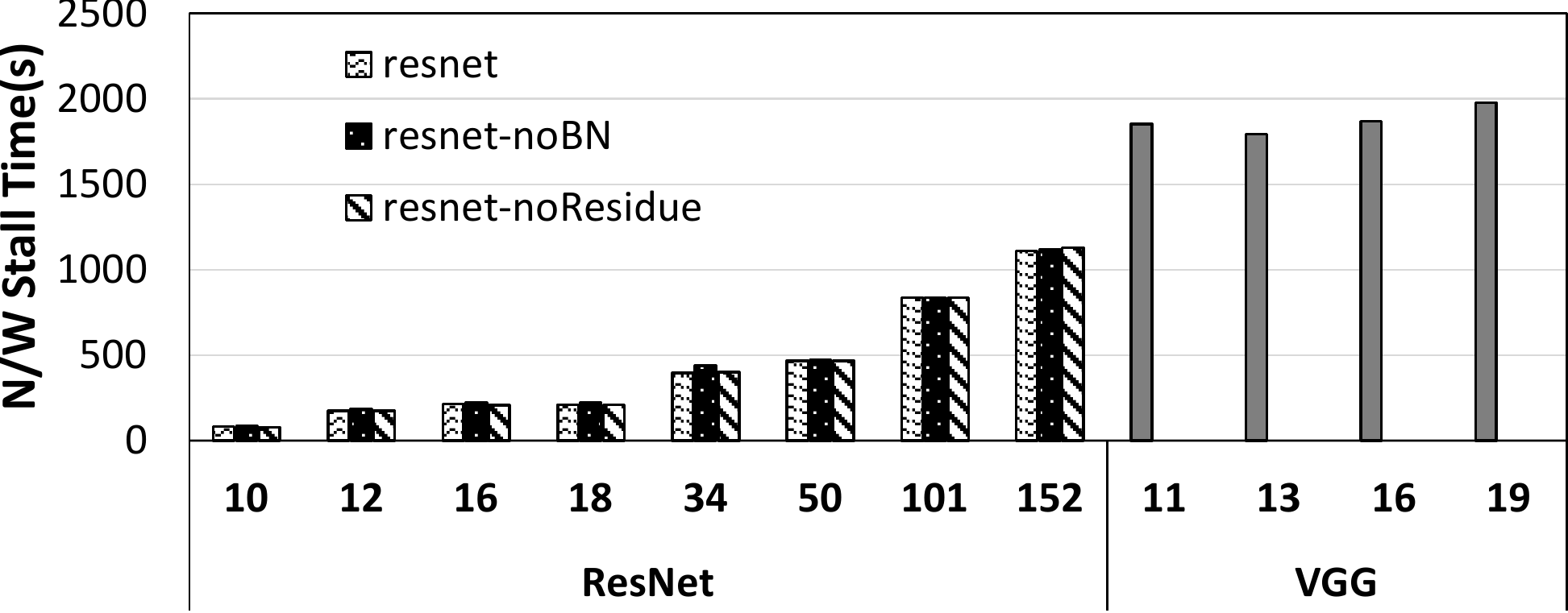} }}
    \subfloat[N/W Stall \%]{{\includegraphics[width=0.49\linewidth]{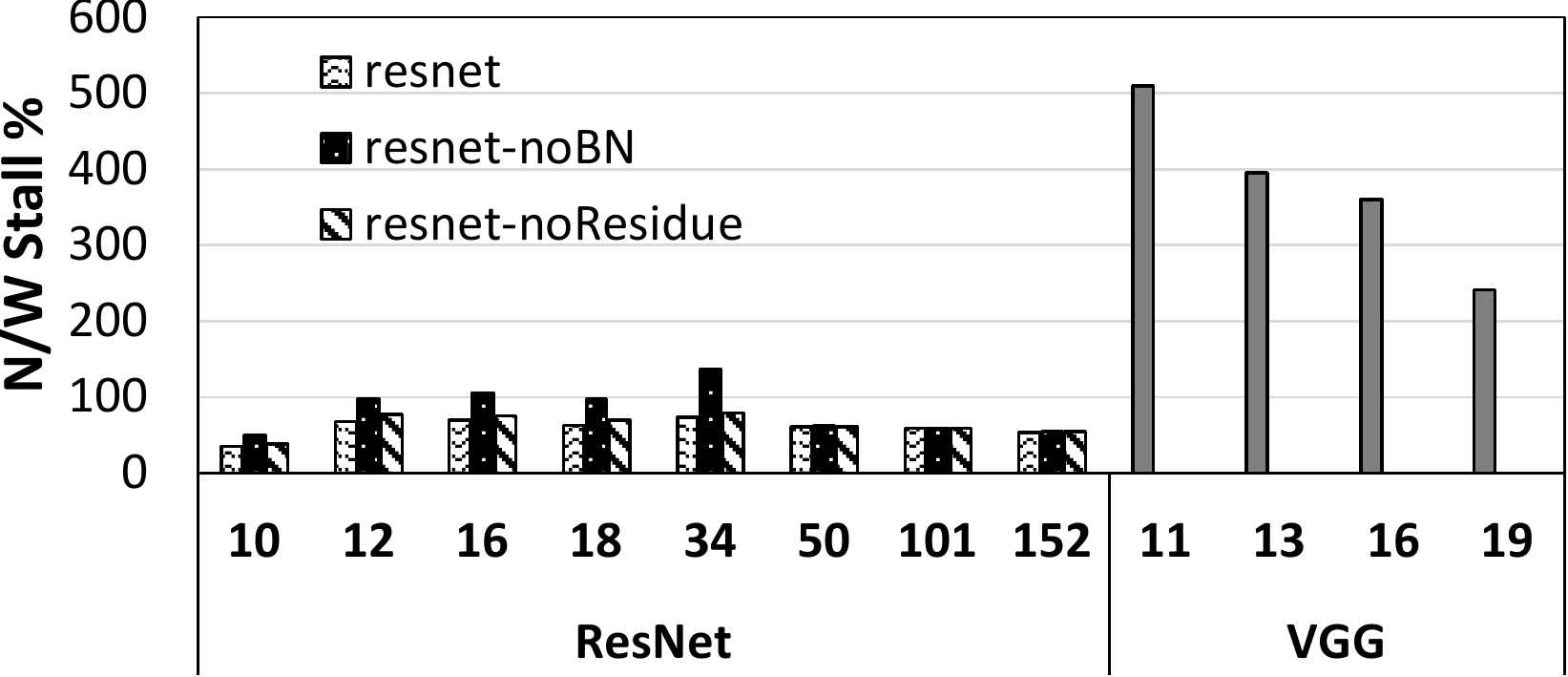} }}
    \vspace{-3mm}
    \caption{Network stall in ResNet and VGG with increasing number of layers. (VGG has higher stalls than ResNet)}
    \label{fig:p3_synthetic_nw_stalls}
    \vspace{-4mm}
\end{figure*}



\subsection{Micro Characterization} \label{subsections: micro_characterization}

In order to verify the generality of our profiling technique for new models, we create synthetic models by altering the model architecture of popular DNNs (namely, ResNet and VGG), to highlight features that can affect stall behavior.
Note that these changes are not meant to improve the DNN model accuracy or training/convergence time.
Rather, they are meant to provide insights that can be leveraged by users to architect DNN models to improve system utilization.
We run all experiments on a p3.16xlarge instance with a batch size of 32 per GPU, repeated thrice with the results averaged across the runs.
A smaller batch size (32) is chosen so as to maximize all-reduce cycles. 
This, in turn, exacerbates communication stalls, thereby, facilitating the analysis of its underlying cause(s).
We begin our discussion by asking two questions: (i) \textbf{Is there a relationship between the number of layers in a model and its communication stalls?}, and (ii) \textbf{How does the number of gradients to be transferred in a model affect communication stalls?} 

\subsubsection{\textbf{Relationship between DNN layers, gradients and communication stalls}}
We answer the above questions by observing the communication stalls of ResNet and VGG with varying number of layers (Figures \ref{fig:p3_synthetic_ic_stalls} and \ref{fig:p3_synthetic_nw_stalls}).
We observe that as the number of layers increases (accompanied by a commensurate increase in the number of gradients), both the interconnect stall and network stall time increases.
This is expected, as there is more data to be transferred with the increase in number of gradients.
However, despite the number of gradients in VGG being far more than that in ResNet (refer table \ref{tab: models_desc}), \emph{VGG is observed to have lower interconnect stall time than ResNet (Figure \ref{fig:p3_synthetic_ic_stalls}) }.
Moreover, we also notice that \emph{the network stall time of VGG is significantly more than ResNet (Figure \ref{fig:p3_synthetic_nw_stalls}).}
These facts lead us to the next question that arises logically: \textbf{Why is the interconnect stall of VGG low and the network stall high when compared to those of ResNet?}
\subsubsection{\textbf{Explaining VGG communication stalls}}

From \cite{pytorch_dist}, we know that distributed PyTorch overlaps communication and computation during the backward pass.
At every layer, there is a synchronization point where communication takes place between the workers.
In case of ResNet, there is a large number of layers and relatively fewer gradients to transfer per layer.
In comparison, VGG has fewer layers, but a larger number of gradients to transfer per layer.
For instance, VGG16 consists of about 134.7 million trainable parameters while ResNet152 consists of only 58.5 million \cite{resnet_vgg_num_parms}.
Therefore, the gradients to transfer per synchronization point is greater in VGG, but the number of times the gradients get transferred is higher in ResNet.
We now explain how this characteristic leads to the interconnect stalls observed in the previous subsection. \par

Suppose VGG has $G_{vgg}$ bytes of gradients and $L_{vgg}$ layers, and ResNet has $G_{res}$ bytes of gradients and $L_{res}$ layers.
Also, let us say that NVLink offers $B_{nv}$ bandwidth with \texttau$_{nv}$ latency.
The time to transfer gradients comprises both latency and data transfer time. 
Define this for VGG and ResNet to respectively be $T_{vgg}$ and $T_{res}$.
Thus, the transfer time using NVlink is given by:
$$
T_{vgg} = \left[\tau_{nv} + \frac{G_{vgg}}{L_{vgg} \times B_{nv}} \right] \times L_{vgg}, ~~
T_{res} = \left[\tau_{nv} + \frac{G_{res}}{L_{res} \times B_{nv}} \right] \times L_{res}
$$

Since NVLink offers very high bandwidth (more than 100 Gbps \cite{li2019evaluating}), and also because both models have a large number of layers, we can assume: 
$\frac{G_{vgg}}{L_{vgg} \times B_{nv}} \ll  \tau_{nv}$ and $\frac{G_{res}}{L_{res} \times B_{nv}} \ll \tau_{nv}$. 
Hence, data transfer time over NVLink is $T_{vgg} \approx \tau_{nv} \times L_{vgg}$ and $T_{res} \approx \tau_{nv} \times L_{res}$.\\\\
Thus, $\boxed{L_{res}$ > $L_{vgg} \implies T_{res}$ > $T_{vgg}}$\\
\\
In other words, the training process experiences increased slowdown due to the larger number of layers to transfer in ResNet (or in any other deep model, for that matter). 
It follows that in the case of VGG, although the data to be transferred is much larger, the time to transfer is nearly zero due to the lower number of layers. The only slowdown we observe here is due to the transfer latency associated with the transfer link/framework.\par

Now, we explain the high network stall time observed for VGG in the previous subsection.
As already explained in Section~\ref{subsection: network_stall}, the collective all-reduce performed across the network-connected instances is throttled by the network link.
Hence, we can assume that the data transfer time is a function of the network link only.
Suppose the network link offers $B_{nw}$ bandwidth with $\tau_{nw}$ latency.
Similarly, the data transfer time over network is:
$$
T_{vgg} = \left[\tau_{nw} + \frac{G_{vgg}}{L_{vgg} \times B_{nw}} \right] \times L_{vgg} ,~~
T_{res} = \left[\tau_{nw} + \frac{G_{res}}{L_{res} \times B_{nw}} \right] \times L_{res}
$$
Since the network link is slow, we can assume:
$\tau_{nw} \ll \frac{G_{vgg}}{B_{nw}}$ and $\tau_{nw} \ll \frac{G_{res}}{B_{nw}}$.
Hence, the data transfer time over network link is: 
$T_{vgg} \approx \frac{G_{vgg}}{B_{nw}}$ and $T_{res} \approx \frac{G_{res}}{B_{nw}}$.\\\\
Thus, $\boxed{G_{vgg}$ > $G_{res} \implies T_{vgg}$ > $T_{res}}$\\
\\
In other words, since the network link is slow, the data transfer is throttled on the transfer time rather than the latency.
Since a much larger volume of gradients needs to be transferred in VGG (in total), the network stall is much higher in VGG than in ResNet.


\subsubsection{\textbf{Impact of model architecture}}
To probe further into the specific aspects of DNN model architecture that impact interconnect stalls, we made two changes to the ResNet model by removing batch normalization (BN) as well as residual networks.
These changes are intended to show the extent to which these layer types impact communication stalls.
From Figures \ref{fig:p3_synthetic_ic_stalls} and \ref{fig:p3_synthetic_nw_stalls}, we notice that removing residual networks has minimal impact on communication overhead.
This is because residual networks do not introduce any new layers and hence do not impact communication.
However, removing BN reduces the number of layers in the model and hence we see lowering in communication stalls as shown in the Figures \ref{fig:p3_synthetic_ic_stalls} and \ref{fig:p3_synthetic_nw_stalls}.

\subsubsection{\textbf{Recommendation}}
From our experiments, we observe that models with very deep networks and fewer gradients are unable to fully exploit the fast NVLink interconnect, whereas shallower networks with large gradient transfers can be throttled on the network link.
Hence, we recommend running shallow networks with large gradients on instances with the best interconnect possible.
However, if the model is very deep with fewer gradients per layer, the models can be run on instances without the best interconnect, such as the p3.8xlarge.
The penalty for running such models on network-connected instances is also minimized.
Note that models can be made more complex through residual connections without affecting communication time whereas removing batch normalization decreases the number of layers affecting communication.

\subsection{Discussion: Network Stall Analysis at Scale} \label{subsections: network_stall_scale}
In this section, we study the
behavior-on-average of network stalls in DDL.
We do this in order to achieve two objectives: (i) to observe how network stalls scale with more instances in the cluster, and (ii) to discuss the challenges of extrapolating the observed network stalls to a cluster with many more instances. 
A potential benefit of such extrapolation is to predict an ideal number of instances to minimize training time of a given DDL workload. We describe this problem as:
\begin{align*}
    \min \left({\sum_{instance ~i=1}^{n} cost_{i}}\right) \quad
    s.t. \quad training\_time < T;
\end{align*}
where the cost of an instance $i$ will (obviously) depend on the total training time.

For experimentation, we used the p3.16xlarge (8 GPU) instance  as it was found to be the most performant through our characterization in the previous section.
All training time used in this analysis is for a single epoch only.
For a simple model, let's say it takes time $T_k$ for running training on 8k GPUs, i.e.,  if just one p3.16xlarge is used then the time taken is $T_1$ which includes both compute and intra-node stalls, but we assume it is dominated by the former.

Now, suppose the training is run over two p3.16xlarge (16 GPUs).
This training time $T_2$ consists of the compute time and network stall overhead of the two instances connected via network link. 
The compute per GPU would be roughly halved as there are twice the number of GPUs and hence half as many mini batches per GPU. 
Let $N_2$ be the network stall associated with the exchange of gradients between the instances in this case. 
Thus, we approximate 
$T_2 \approx T_1/2+N_2.$
Similarly, for an $n$ instance cluster, training time $T_n$ with network overhead $N_n$ is:
$$T_n \approx T_1/n + N_n$$

Regarding the network stalls, $N$,
let's say the total volume of the gradients computed per GPU after backward pass is $G$ bytes (which is irrespective of the number of GPUs).
We now argue for the following 
highly idealized approximation (refer Section~\ref{subsection: ddl_background}):
$$N_n \approx (\tau + \frac{2G/n}{B}) \times (n-1),$$
where $\tau$ is the (average) pairwise transfer latency between instances (which may also include the effects of time-varying network congestion), $B$ is the bandwidth (transmission rate in the instance SLA), and the term $n-1$ represents the maximum number of hops when the instances are arranged in a ring (consistent with an idealized ``all-reduce" topology \cite{all-reduce-towards-data-science}). Note that this approximation can be adapted for cases where the topology is different (e.g., inverted tree) or when gradients are averaged at the instance level prior to transmission to other instances (i.e., local synchronization). In a practical cloud setting, the placement of instances by the cloud provider was not considered above and is an important complication to extrapolating network stalls, e.g., on the same server, rack, LAN or in different availability zones.  (Here, one may reject a newly acquired instance as the cluster autoscales if its associated network delay is measured as very large.) In some cases, a user can pay more to have their instances proximal to one another.

Combining the previous two displays, we get:
$$T_n = T_1/n + (\tau + \frac{2G/n}{B}) \times (n-1),$$
which is a convex function of $n$ (where
the third term is roughly constant).
For this idealized model, if $\sqrt{T_1/\tau}>1$ then there is an optimal number of instances,
 $\lfloor{\sqrt{T_1/\tau}}\rfloor$ or
  $\lceil{\sqrt{T_1/\tau}}\rceil$; otherwise, a single instance is optimal (model with low compute requirement).
Note that this simple model can be based on the profiling seen in the previous section.
  
Though highly idealized, the previous display is an example of a more general rule-of-thumb well known in parallel computing: as the number of instances $n$ grows, the communication component grows ($n\tau$) while the computation component diminishes
  ($T_1/n$), and the former will eventually dominate.

\begin{figure}
\centering
    {\includegraphics[width=0.4\textwidth]{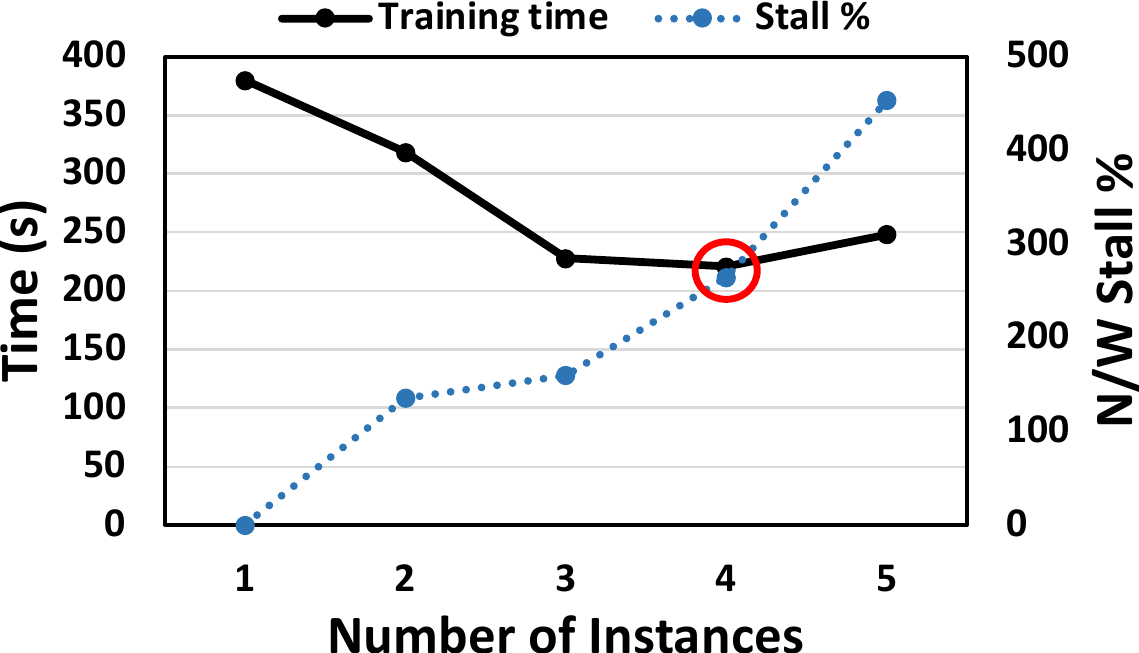} }
\caption{Training time and network stall \% scaling for ResNet50 with increasing instances}
\label{fig:nw_scale}
\end{figure}

To validate this theory, we run multi-node DDL with ResNet50.
We scale the number of instances from one to five and observe the training time in Figure \ref{fig:nw_scale}.
As can be observed, training time first decreases up till four instances and then starts increasing beyond that point.
This is because at the point of about four instances, communication starts eclipsing the compute.
As such, the lowest training time a user can expect to train ResNet50 is by using four instances considering network remains the same.
We also notice from the figure that the network stall keeps on increasing as instances are added.

Note that experimentally, we found that execution time may increase significantly when scaling from one instance to two for large models (e.g., our BERT and VGG case studies). This is due to the need to exchange a large volume of gradient data. 
Unfortunately, we couldn't scale beyond 5 instances due to resource unavailability in AWS.



In future work, we will study how to quickly ascertain the throughput and transfer delays of a given cluster topology as scheduled by the cloud provider, and how to coordinate gradient sharing to minimize the impact of networks stalls on overall execution time \cite{huang2021communication,rashidi2022themis}.\par

\vspace{-3mm}
\section{Related Work} \label{sections: relatedwork}
This section discusses the related work pertaining to DL characterization, in addition to covering specific works in DDL that focus on reducing communication overheads and optimizing costs in the public cloud systems.


\textbf{Characterizing Deep Learning.~}
There is a wealth of existing work in the area of deep learning characterization \cite{awan2017depth,awan2019scalable,jain2019performance,mojumder2018profiling,xia2018characterizing,wang2019characterizing,li2020characterizing,samajdar2020systematic,hu2021characterization,mohan2021,weng2022mlaas,liu2019performance} but none that conducts a stall analysis on public cloud GPU instances.
In \cite{weng2022mlaas}, the authors analyze 2 months of MLaaS production data in Alibaba cloud \cite{alibaba_cloud} and provide insights into its cluster scheduling and low GPU utilization.
\cite{wang2019characterizing} characterizes deep learning workloads on the Alibaba-PAI platform \cite{alibaba_pai} and \cite{li2020characterizing} builds a measurement and prediction framework to train convolution neural networks with transient VMs.


\textbf{Optimizing DDL communication.~}
\cite{pytorch_dist,rashidi2021enabling} aim to reduce communication time by overlapping it with computation. 
\cite{zhang2015deep,lian2018asynchronous,mosharraf_parm_server,ho_more_effective_distributed_ml,goyal2017accurate,jeff_dean_Large_Scale_Distributed_Deep_Networks,rashidi2022themis,dong2020eflops,nccl,hashemi2019tictac,jayarajan2019priority,lin2017deep,wang2018blink,zhang2017poseidon} are proposed algorithmic changes to reduce communication time in DDL. Note that these  works do not optimize the underlying system being used. 
\cite{rashidi2020scalable,rashidi2020astra} are simulation platforms to measure and build new communication-based algorithms for DDL. However, they are agnostic about the underlying hardware-induced delays.

\textbf{Cost optimization in the public cloud.~}
Public cloud providers offer cost management tools
that enable tenants to monitor and
optimize their cloud-spend \cite{cloudspend}
e.g., \cite{azure-cost-mgmt}.
Prior work such as, 
\cite{costless,mark-atc19} 
explicitly
focus on reducing costs of migrating and running
"generic" workloads on public cloud 
(including DNN inference), i.e.,
not specifically deep learning.
Some of this work is described
as efficient resource management, which may not 
explicitly discuss monetary costs, 
but implicitly does. For example, the use of serverless
functions (e.g., Amazon Lambda) for 
(not stateful) microservice-based
applications
is a cost-based
choice because serverless functions have
a finer pricing granularity, shorter
(warm start) spin-up times, and
less management overhead compared to virtual machines,
e.g., \cite{Iaas-compare} (even though their
cost per unit-resource per unit time is higher). 
Note that DDL is obviously quite stateful. Also,
 prior work, e.g., \cite{Seer19b}, has 
applied deep learning to the problem of efficient
resource management.

\begin{table}[]
\resizebox{1\linewidth}{!}{
\begin{tabular}{|l|l|}
\hline
\multicolumn{1}{|c|}{\textbf{Observation (Figure\#)}}                                            & \multicolumn{1}{c|}{\textbf{Insight (Section\#)}}                                                   \\ \hline
High disk stall in p2.16xlarge (\ref{fig:p2_cpu_disk_stall}(b))                   & Disk performance insufficient (\ref{subsections: p2_evaluation})                \\ \hline
High I/C stall in p2.16xlarge (\ref{fig:p2_interconnect_stall})                   & PCIe bus bandwidth contention (\ref{subsections: p2_evaluation})                \\ \hline
High I/C stall in p3.8xlarge (\ref{fig:p3_interconnect_stall_small})             & Sub-optimal interconnect allocation (\ref{subsections: p3_eval})                \\ \hline
Low cost optimality of p3.24xlarge (\ref{fig:p3_train_time_cost_large}(b))        & Interconnect same as 16xlarge (\ref{subsections: p3_eval})                      \\ \hline
High cost optimality of p3.8xlarge (\ref{fig:p3_train_time_cost_large}(b))        & Higher disk stall in larger instances (\ref{subsections: p3_eval})                \\ \hline
High Network stalls (\ref{fig:network_overhead})                                  & Network QoS insufficient for DDL (\ref{subsections: p3_eval})                   \\ \hline
Cost optimality of P3 better than P2 (\ref{fig:p2_p3_cost_comparison}(b))           & P3 suffers from lower stalls than P2 (\ref{subsections: p2_vs_p3})                       \\ \hline
Higher cost optimality of small models in P2 (\ref{fig:p2_p3_cost_comparison}(b)) & Inability to exploit P3 resources (\ref{subsections: p2_vs_p3})                      \\ \hline
Low I/C stalls of fat \& shallow models (\ref{fig:p3_synthetic_ic_stalls})               & NVlink offers ample bandwidth (\ref{subsections: micro_characterization})       \\ \hline
High network stalls of large models (\ref{fig:p3_synthetic_nw_stalls})            & Network throttles gradient transfer (\ref{subsections: micro_characterization}) \\ \hline
Training time scales as a convex function (\ref{fig:nw_scale})                    & Network stall dominates with increasing instances (\ref{subsections: network_stall_scale})           \\ \hline
\end{tabular} }
\caption{Key observations and insights of this study}
\label{tab: key_insights}
\vspace{-8mm}
\end{table}




\vspace{-3mm}
\section{Concluding Remarks} \label{sections: summary}
We introduced a DDL stall profiler \name and presented novel methodologies to measure communication stalls. Using the profiler, we extensively characterized  public cloud GPU instances for the various stalls they experience when running popular DNNs. We found communication stall to be the major bottleneck in DDL training and that AWS instances suffer heavily from communication stalls. The observed interconnect stalls and network stalls were up to 90\% of single GPU and 500\% of single instance training time respectively. The reasoning for these high stalls were stated to be severe bandwidth contention when using the PCIe bus, sub-optimal resource allocation when using the NVLink and low AWS network bandwidth. The high stall numbers directly translated to higher training cost compared to instances with lower communication stalls. Further, we explained analytically how certain DNN model architecture features (layers and gradient size) drive communication stall behavior, and how users can modify their model architecture to lower communication stalls. And, finally, we discussed how network stall scales through modelling and empirical observations. 
Note that while our experiments were run on AWS, the methods described herein are generic and can be applied on any public cloud. We refer readers to Table~\ref{tab: key_insights} for a summary of the key findings.

\bibliographystyle{plain}
\bibliography{references,cloud,scheduling,DDoS,dnn,sysml}

\begin{thebibliography}{10}

\bibitem{tensorflow}
Mart\'{\i}n Abadi, Paul Barham, Jianmin Chen, Zhifeng Chen, Andy Davis, Jeffrey
  Dean, Matthieu Devin, Sanjay Ghemawat, Geoffrey Irving, Michael Isard,
  Manjunath Kudlur, Josh Levenberg, Rajat Monga, Sherry Moore, Derek~G. Murray,
  Benoit Steiner, Paul Tucker, Vijay Vasudevan, Pete Warden, Martin Wicke, Yuan
  Yu, and Xiaoqiang Zheng.
\newblock {TensorFlow: A System for Large-scale Machine Learning}.
\newblock In {\em Proc. USENIX OSDI}, 2016.

\bibitem{alibaba_cloud}
{Alibaba Cloud}.
\newblock \url{https://www.alibabacloud.com/}, Accessed: 2022.06.15.

\bibitem{awan2019scalable}
Ammar~Ahmad Awan, Jereon B{\'e}dorf, Ching-Hsiang Chu, Hari Subramoni, and
  Dhabaleswar~K Panda.
\newblock {Scalable distributed DNN training using TensorFlow and CUDA-aware
  MPI: Characterization, designs, and performance evaluation}.
\newblock In {\em Proc. IEEE/ACM CCGRID}, 2019.

\bibitem{awan2017depth}
Ammar~Ahmad Awan, Hari Subramoni, and Dhabaleswar~K Panda.
\newblock {An in-depth performance characterization of CPU-and GPU-based DNN
  training on modern architectures}.
\newblock In {\em Proceedings of the Machine Learning on HPC Environments},
  2017.

\bibitem{aws}
{Amazon Web Services (AWS)}.
\newblock \url{https://aws.amazon.com/}, 2022.06.08.

\bibitem{aws_nvidia}
{AWS NVIDIA GPU instances}.
\newblock \url{https://aws.amazon.com/nvidia/}, Accessed: 2022.06.08.

\bibitem{AWS-HPC}
{AWS HPC}.
\newblock https://aws.amazon.com/hpc/, Accessed: 2022-06-08.

\bibitem{Azure-HPC}
{Azure HPC}.
\newblock
  \url{https://azure.microsoft.com/en-us/solutions/high-performance-computing/#overview},
  Accessed: 2022-06-08.

\bibitem{alibaba_pai}
{Alibaba PAI}.
\newblock \url{https://github.com/AlibabaPAI}, Accessed: 2022.06.15.

\bibitem{dawnBench}
{DawnBench}.
\newblock \url{https://dawn.cs.stanford.edu/benchmark/}, Accessed: 2022-06-08.

\bibitem{model_pll}
Jeffrey Dean, Greg Corrado, Rajat Monga, Kai Chen, Matthieu Devin, Mark Mao,
  Marc\textquotesingle~aurelio Ranzato, Andrew Senior, Paul Tucker, Ke~Yang,
  Quoc Le, and Andrew Ng.
\newblock Large scale distributed deep networks.
\newblock In F.~Pereira, C.J. Burges, L.~Bottou, and K.Q. Weinberger, editors,
  {\em Advances in Neural Information Processing Systems}, volume~25. Curran
  Associates, Inc., 2012.

\bibitem{jeff_dean_Large_Scale_Distributed_Deep_Networks}
Jeffrey Dean, Greg Corrado, Rajat Monga, Kai Chen, Matthieu Devin, Mark Mao,
  Marc\textquotesingle~aurelio Ranzato, Andrew Senior, Paul Tucker, Ke~Yang,
  Quoc Le, and Andrew Ng.
\newblock Large scale distributed deep networks.
\newblock In {\em Proc. Advances in Neural Information Processing Systems},
  volume~25, 2012.

\bibitem{bert}
Jacob Devlin, Ming{-}Wei Chang, Kenton Lee, and Kristina Toutanova.
\newblock {BERT:} pre-training of deep bidirectional transformers for language
  understanding.
\newblock In Jill Burstein, Christy Doran, and Thamar Solorio, editors, {\em
  Proceedings of the 2019 Conference of the North American Chapter of the
  Association for Computational Linguistics: Human Language Technologies,
  {NAACL-HLT} 2019, Minneapolis, MN, USA, June 2-7, 2019, Volume 1 (Long and
  Short Papers)}, pages 4171--4186. Association for Computational Linguistics,
  2019.

\bibitem{googleBERT}
Jacob Devlin, Ming-Wei Chang, Kenton Lee, and Kristina Toutanova.
\newblock {BERT: Pre-training of Deep Bidirectional Transformers for Language
  Understanding}.
\newblock https://arxiv.org/abs/1810.04805, 24 May 2019.

\bibitem{dong2020eflops}
Jianbo Dong, Zheng Cao, Tao Zhang, Jianxi Ye, Shaochuang Wang, Fei Feng,
  Li~Zhao, Xiaoyong Liu, Liuyihan Song, Liwei Peng, et~al.
\newblock Eflops: Algorithm and system co-design for a high performance
  distributed training platform.
\newblock In {\em 2020 IEEE International Symposium on High Performance
  Computer Architecture (HPCA)}, pages 610--622. IEEE, 2020.

\bibitem{ds-analyzer}
{DS-Analyzer}.
\newblock \url{https://github.com/msr-fiddle/DS-Analyzer}, Accessed:
  2022.06.08.

\bibitem{costless}
Tarek {Elgamal}, A.~{Sandur}, K.~{Nahrstedt}, and G.~{Agha}.
\newblock {Costless: Optimizing Cost of Serverless Computing through Function
  Fusion and Placement}.
\newblock In {\em IEEE/ACM Symposium on Edge Computing (SEC)}, pages 300--312,
  Seattle, WA, USA, Oct. 25-27, 2018.

\bibitem{npu0}
Jeremy Fowers and al.
\newblock {A Configurable Cloud-Scale DNN Processor for Real-Time AI}.
\newblock In {\em Proc. ACM ISCA}, June 2018.

\bibitem{Seer19b}
Yu~Gan, Yanqi Zhang, Kelvin Hu, Dailun Cheng, Yuan He, Meghna Pancholi, and
  Christina Delimitrou.
\newblock {Leveraging Deep Learning to Improve Performance Predictability in
  Cloud Microservices with Seer}.
\newblock {\em ACM SIGOPS Operating Systems Review}, 53(1), July 2019.

\bibitem{gcloud}
{Google Cloud : Cloud Computing Services}.
\newblock \url{https://cloud.google.com/}, Accessed: 2022.06.15.

\bibitem{goyal2017accurate}
Priya Goyal, Piotr Doll{\'a}r, Ross Girshick, Pieter Noordhuis, Lukasz
  Wesolowski, Aapo Kyrola, Andrew Tulloch, Yangqing Jia, and Kaiming He.
\newblock {Accurate, large minibatch SGD: Training ImageNet in 1 hour}.
\newblock {\em arXiv preprint arXiv:1706.02677}, 2017.

\bibitem{hashemi2019tictac}
Sayed~Hadi Hashemi, Sangeetha Abdu~Jyothi, and Roy Campbell.
\newblock Tictac: Accelerating distributed deep learning with communication
  scheduling.
\newblock {\em Proceedings of Machine Learning and Systems}, 1:418--430, 2019.

\bibitem{resnet}
Kaiming He, Xiangyu Zhang, Shaoqing Ren, and Jian Sun.
\newblock Identity mappings in deep residual networks.
\newblock In Bastian Leibe, Jiri Matas, Nicu Sebe, and Max Welling, editors,
  {\em Computer Vision -- ECCV 2016}, 2016.

\bibitem{ho_more_effective_distributed_ml}
Qirong Ho, James Cipar, Henggang Cui, Jin~Kyu Kim, Seunghak Lee, Phillip~B.
  Gibbons, Garth~A. Gibson, Gregory~R. Ganger, and Eric~P. Xing.
\newblock More effective distributed ml via a stale synchronous parallel
  parameter server.
\newblock In {\em Proc. NIPS}, 2013.

\bibitem{hu2021characterization}
Qinghao Hu, Peng Sun, Shengen Yan, Yonggang Wen, and Tianwei Zhang.
\newblock {Characterization and prediction of deep learning workloads in
  large-scale GPU datacenters}.
\newblock In {\em Proceedings of the International Conference for High
  Performance Computing, Networking, Storage and Analysis}, pages 1--15, 2021.

\bibitem{huang2021communication}
Jiayi Huang, Pritam Majumder, Sungkeun Kim, Abdullah Muzahid, Ki~Hwan Yum, and
  Eun~Jung Kim.
\newblock Communication algorithm-architecture co-design for distributed deep
  learning.
\newblock In {\em 2021 ACM/IEEE 48th Annual International Symposium on Computer
  Architecture (ISCA)}, pages 181--194. IEEE, 2021.

\bibitem{DAD}
Yu~Huang and Yue Chen.
\newblock Autonomous driving with deep learning: A survey of state-of-art
  technologies.
\newblock \url{https://arxiv.org/abs/2006.06091}, 2020.

\bibitem{squeezenet}
Forrest~N. Iandola, Song Han, Matthew~W. Moskewicz, Khalid Ashraf, William~J.
  Dally, and Kurt Keutzer.
\newblock {SqueezeNet: AlexNet-level accuracy with 50x fewer parameters and
  $<$0.5MB model size}.
\newblock \url{https://arxiv.org/abs/1602.07360}, 2016.

\bibitem{imagenet_ILSVRC2012}
{Imagenet 1K}.
\newblock \url{https://image-net.org/challenges/LSVRC/2012/}.

\bibitem{imagenet1k}
{ImageNet Large Scale Visual Recognition Challenge 2012 (ILSVRC2012)}.
\newblock \url{https://www.image-net.org/challenges/LSVRC/2012/}, Accessed:
  2022.10.18.

\bibitem{jain2019performance}
Arpan Jain, Ammar~Ahmad Awan, Quentin Anthony, Hari Subramoni, and Dhableswar
  K~DK Panda.
\newblock Performance characterization of dnn training using tensorflow and
  pytorch on modern clusters.
\newblock In {\em 2019 IEEE International Conference on Cluster Computing
  (CLUSTER)}, pages 1--11. IEEE, 2019.

\bibitem{jayarajan2019priority}
Anand Jayarajan, Jinliang Wei, Garth Gibson, Alexandra Fedorova, and Gennady
  Pekhimenko.
\newblock {Priority-based parameter propagation for distributed DNN training}.
\newblock {\em Proceedings of Machine Learning and Systems}, 1:132--145, 2019.

\bibitem{attention-layer}
D.~Jurafsky and J.H. Martin.
\newblock {Speech and Language Processing (3rd ed. draft)}.
\newblock \url{https://web.stanford.edu/~jurafsky/slp3/}, Dec 29, 2021.

\bibitem{keskar2016large}
Nitish~Shirish Keskar, Dheevatsa Mudigere, Jorge Nocedal, Mikhail Smelyanskiy,
  and Ping Tak~Peter Tang.
\newblock On large-batch training for deep learning: Generalization gap and
  sharp minima.
\newblock {\em arXiv preprint arXiv:1609.04836}, 2016.

\bibitem{p3_interconnect}
Rashika Kheria, Purna Sanyal, Sr. James~Jeun, and Amr Ragab.
\newblock {Optimizing deep learning on P3 and P3dn with EFA }.
\newblock
  \url{https://aws.amazon.com/blogs/compute/optimizing-deep-learning-on-p3-and-p3dn-with-efa//},
  Accessed: 2022.06.08.

\bibitem{ko2021depth}
Yunyong Ko, Kibong Choi, Jiwon Seo, and Sang-Wook Kim.
\newblock An in-depth analysis of distributed training of deep neural networks.
\newblock In {\em 2021 IEEE International Parallel and Distributed Processing
  Symposium (IPDPS)}, pages 994--1003. IEEE, 2021.

\bibitem{alexnet}
Alex Krizhevsky, Ilya Sutskever, and Geoffrey~E Hinton.
\newblock Imagenet classification with deep convolutional neural networks.
\newblock In {\em Advances in Neural Information Processing Systems},
  volume~25, 2012.

\bibitem{all-reduce-towards-data-science}
Edir~Garcia Lazo.
\newblock {Visual intuition on ring-Allreduce for distributed Deep Learning}.
\newblock
  \url{https://towardsdatascience.com/visual-intuition-on-ring-allreduce-for-distributed-deep-learning-d1f34b4911da},
  Accessed: 2022.10.18.

\bibitem{resnet_vgg_num_parms}
Mei Leong, Dilip Prasad, Yong~Tsui Lee, and Feng Lin.
\newblock Semi-cnn architecture for effective spatio-temporal learning in
  action recognition.
\newblock {\em Applied Sciences}, 10:557, 01 2020.

\bibitem{li2019evaluating}
Ang Li, Shuaiwen~Leon Song, Jieyang Chen, Jiajia Li, Xu~Liu, Nathan~R Tallent,
  and Kevin~J Barker.
\newblock {Evaluating modern GPU interconnect: PCIe, NVLink, NV-SLI, NVSwitch
  and GPUDirect}.
\newblock {\em IEEE Transactions on Parallel and Distributed Systems},
  31(1):94--110, 2019.

\bibitem{mosharraf_parm_server}
Mu~Li, David~G. Andersen, Jun~Woo Park, Alexander~J. Smola, Amr Ahmed, Vanja
  Josifovski, James Long, Eugene~J. Shekita, and Bor-Yiing Su.
\newblock Scaling distributed machine learning with the parameter server.
\newblock In {\em Proc. USENIX OSDI}, 2014.

\bibitem{pytorch_dist}
Shen Li, Yanli Zhao, Rohan Varma, Omkar Salpekar, Pieter Noordhuis, Teng Li,
  Adam Paszke, Jeff Smith, Brian Vaughan, Pritam Damania, and Soumith Chintala.
\newblock {PyTorch Distributed: Experiences on Accelerating Data Parallel
  Training}.
\newblock {\em Proc. VLDB Endow.}, 13(12):3005–3018, aug 2020.

\bibitem{li2020characterizing}
Shijian Li, Robert~J Walls, and Tian Guo.
\newblock {Characterizing and modeling distributed training with transient
  cloud GPU servers}.
\newblock In {\em Proc. IEEE 40th International Conference on Distributed
  Computing Systems (ICDCS)}, pages 943--953, 2020.

\bibitem{lian2018asynchronous}
Xiangru Lian, Wei Zhang, Ce~Zhang, and Ji~Liu.
\newblock Asynchronous decentralized parallel stochastic gradient descent.
\newblock In {\em International Conference on Machine Learning}, pages
  3043--3052. PMLR, 2018.

\bibitem{lin2017deep}
Yujun Lin, Song Han, Huizi Mao, Yu~Wang, and William~J Dally.
\newblock Deep gradient compression: Reducing the communication bandwidth for
  distributed training.
\newblock {\em arXiv preprint arXiv:1712.01887}, 2017.

\bibitem{liu2019performance}
Jie Liu, Jiawen Liu, Wan Du, and Dong Li.
\newblock Performance analysis and characterization of training deep learning
  models on mobile device.
\newblock In {\em 2019 IEEE 25th International Conference on Parallel and
  Distributed Systems (ICPADS)}, pages 506--515. IEEE, 2019.

\bibitem{shufflenet_v2}
Ningning Ma, Xiangyu Zhang, Hai-Tao Zheng, and Jian Sun.
\newblock Shufflenet v2: Practical guidelines for efficient cnn architecture
  design.
\newblock In {\em Proceedings of the European Conference on Computer Vision
  (ECCV)}, September 2018.

\bibitem{mohan2021}
Jayashree Mohan, Amar Phanishayee, Ashish Raniwala, and Vijay Chidambaram.
\newblock Analyzing and mitigating data stalls in dnn training.
\newblock https://arxiv.org/abs/2007.06775, 2021.

\bibitem{mojumder2018profiling}
Saiful~A Mojumder, Marcia~S Louis, Yifan Sun, Amir~Kavyan Ziabari, Jos{\'e}~L
  Abell{\'a}n, John Kim, David Kaeli, and Ajay Joshi.
\newblock {Profiling DNN workloads on a Volta-based DGX-1 system}.
\newblock In {\em Proc. IEEE International Symposium on Workload
  Characterization (IISWC)}, pages 122--133. IEEE, 2018.

\bibitem{azure}
{Microsoft Azure}.
\newblock \url{https://azure.microsoft.com/}, 2022.06.08.

\bibitem{msr-fiddle}
{Project Fiddle}.
\newblock \url{https://aka.ms/msr-fiddle}, Accessed: 2022-06-08.

\bibitem{pipedream}
Deepak Narayanan, Aaron Harlap, Amar Phanishayee, Vivek Seshadri, Nikhil~R.
  Devanur, Gregory~R. Ganger, Phillip~B. Gibbons, and Matei Zaharia.
\newblock {PipeDream: Generalized Pipeline Parallelism for DNN Training}.
\newblock In {\em Proc. SOSP}, 2019.

\bibitem{nccl}
{NCCL}.
\newblock \url{https://developer.nvidia.com/nccl}, Accessed: 2022.06.15.

\bibitem{nvidia_pyTorch_egs}
{NVIDIA Deep Learning Examples for Tensor Cores }.
\newblock \url{https://github.com/NVIDIA/DeepLearningExamples}, Accessed:
  2022.06.08.

\bibitem{azure-cost-mgmt}
{Optimize your Azure costs}.
\newblock \url{https://azure.microsoft.com/en-us/overview/cost-optimization/}.

\bibitem{image-transformer}
N.J. Parmar, A.~Vaswani, J.~Uszkoreit, L.~Kaiser, N.~Shazeer, A.~Ku, and
  D.~Tran.
\newblock {Image Transformer}.
\newblock In {\em Proc. International Conference on Machine Learning (ICML)},
  2018.

\bibitem{squad}
Pranav Rajpurkar, Jian Zhang, Konstantin Lopyrev, and Percy Liang.
\newblock Squad: 100,000+ questions for machine comprehension of text, 2016.

\bibitem{rashidi2021enabling}
Saeed Rashidi, Matthew Denton, Srinivas Sridharan, Sudarshan Srinivasan,
  Amoghavarsha Suresh, Jade Nie, and Tushar Krishna.
\newblock Enabling compute-communication overlap in distributed deep learning
  training platforms.
\newblock In {\em Proc. ACM/IEEE ISCA}, pages 540--553, 2021.

\bibitem{rashidi2020scalable}
Saeed Rashidi, Pallavi Shurpali, Srinivas Sridharan, Naader Hassani, Dheevatsa
  Mudigere, Krishnakumar Nair, Misha Smelyanski, and Tushar Krishna.
\newblock {Scalable distributed training of recommendation models: An
  Astra-Sim+ ns3 case-study with TCP/IP transport}.
\newblock In {\em 2020 IEEE Symposium on High-Performance Interconnects
  (HOTI)}, pages 33--42. IEEE, 2020.

\bibitem{rashidi2020astra}
Saeed Rashidi, Srinivas Sridharan, Sudarshan Srinivasan, and Tushar Krishna.
\newblock {Astra-Sim: Enabling SW/HW co-design exploration for distributed DL
  training platforms}.
\newblock In {\em Proc. IEEE International Symposium on Performance Analysis of
  Systems and Software (ISPASS)}, pages 81--92, 2020.

\bibitem{rashidi2022themis}
Saeed Rashidi, William Won, Sudarshan Srinivasan, Srinivas Sridharan, and
  Tushar Krishna.
\newblock {Themis: A network bandwidth-aware collective scheduling policy for
  distributed training of DL models}.
\newblock In {\em Proc.ACM/IEEE ISCA}, pages 581--596, 2022.

\bibitem{samajdar2020systematic}
Ananda Samajdar, Jan~Moritz Joseph, Yuhao Zhu, Paul Whatmough, Matthew Mattina,
  and Tushar Krishna.
\newblock {A systematic methodology for characterizing scalability of DNN
  accelerators using scale-sim}.
\newblock In {\em Proc. IEEE International Symposium on Performance Analysis of
  Systems and Software (ISPASS)}, pages 58--68, 2020.

\bibitem{mobilenet}
Mark Sandler, Andrew Howard, Menglong Zhu, Andrey Zhmoginov, and Liang-Chieh
  Chen.
\newblock Mobilenetv2: Inverted residuals and linear bottlenecks.
\newblock In {\em 2018 IEEE/CVF Conference on Computer Vision and Pattern
  Recognition}, pages 4510--4520, 2018.

\bibitem{vgg}
Karen Simonyan and Andrew Zisserman.
\newblock Very deep convolutional networks for large-scale image recognition.
\newblock In {\em Proc. ICLR}, San Diego, CA, May 7-9, 2015.

\bibitem{Iaas-compare}
M.~Villamizar, O.~Garcés, L.~Ochoa, H.~Castro, L.~Salamanca, M.~Verano,
  R.~Casallas, S.~Gil, C.~Valencia, A.~Zambrano, and M.~Lang.
\newblock {Infrastructure Cost Comparison of Running Web Applications in the
  Cloud Using AWS Lambda and Monolithic and Microservice Architectures}.
\newblock In {\em {Proc. IEEE/ACM CCGrid}}, 2016.

\bibitem{wang2018blink}
Guanhua Wang, Amar Phanishayee, Shivaram Venkataraman, and Ion Stoicat.
\newblock Blink: A fast nvlink-based collective communication library.
\newblock In {\em Proc. Conf. Syst. Mach. Learn}, 2018.

\bibitem{wang2019characterizing}
Mengdi Wang, Chen Meng, Guoping Long, Chuan Wu, Jun Yang, Wei Lin, and Yangqing
  Jia.
\newblock {Characterizing deep learning training workloads on Alibaba-PAI}.
\newblock In {\em Proc. IEEE Int'l Symp. on Workload Characterization (IISWC)},
  pages 189--202, 2019.

\bibitem{tpu0}
Yu~Wang, Gu{-}Yeon Wei, and David Brooks.
\newblock {Benchmarking TPU, GPU, and {CPU} Platforms for Deep Learning}.
\newblock \url{http://arxiv.org/abs/1907.10701}, 2019.

\bibitem{weng2022mlaas}
Qizhen Weng, Wencong Xiao, Yinghao Yu, Wei Wang, Cheng Wang, Jian He, Yong Li,
  Liping Zhang, Wei Lin, and Yu~Ding.
\newblock {MLaaS in the Wild: Workload Analysis and Scheduling in
  $\{$Large-Scale$\}$ Heterogeneous GPU Clusters}.
\newblock In {\em Proc. USENIX NSDI}, pages 945--960, 2022.

\bibitem{cloudspend}
B.~Whittle.
\newblock {Navigating Economic Uncertainty in 2020: Cutting Cloud Costs}.
\newblock
  \url{https://www.apptio.com/blog/navigating-economic-uncertainty-in-2020-cost-cutting-cloud/},
  April 6, 2020.

\bibitem{xia2018characterizing}
Chunwei Xia, Jiacheng Zhao, Huimin Cui, and Xiaobing Feng.
\newblock Characterizing dnn models for edge-cloud computing.
\newblock In {\em 2018 IEEE International Symposium on Workload
  Characterization (IISWC)}, pages 82--83, 2018.

\bibitem{mark-atc19}
Chengliang Zhang, Minchen Yu, Wei Wang, and Feng Yan.
\newblock {MArk: Exploiting Cloud Services for Cost-Effective, SLO-Aware
  Machine Learning Inference Serving}.
\newblock In {\em Proc. USENIX ATC}, Renton, WA, 2019.

\bibitem{zhang2017poseidon}
Hao Zhang, Zeyu Zheng, Shizhen Xu, Wei Dai, Qirong Ho, Xiaodan Liang, Zhiting
  Hu, Jinliang Wei, Pengtao Xie, and Eric~P Xing.
\newblock {Poseidon: An efficient communication architecture for distributed
  deep learning on GPU clusters}.
\newblock In {\em Proc. USENIX ATC}, 2017.

\bibitem{zhang2015deep}
Sixin Zhang, Anna~E Choromanska, and Yann LeCun.
\newblock {Deep learning with elastic averaging SGD}.
\newblock {\em Proc. Advances in neural information processing systems}, 28,
  2015.

\end{thebibliography}




\end{document}